\documentclass[letterpaper, 10 pt, conference]{ieeeconf}  % Comment this line out if you need a4paper

\IEEEoverridecommandlockouts                              % This command is only needed if 
\usepackage{graphics} % for pdf, bitmapped graphics files
\usepackage{epsfig} % for postscript graphics files
\usepackage{times} % assumes new font selection scheme installed
\usepackage{amsmath} % assumes amsmath package installed
\usepackage{amssymb}  % assumes amsmath package installed
\usepackage{bbm}
\usepackage{booktabs}
\usepackage{xcolor}
\usepackage{float}
\usepackage{cite}

\usepackage{graphicx}
\usepackage{float}
\usepackage{subcaption}
\usepackage{multirow}
\usepackage{flushend}
\usepackage{algorithm}
\usepackage{algpseudocode}

\newcommand{\KL}{\mathrm{KL}}
\newcommand{\HLG}{\mathrm{HLG}}
\newcommand{\method}{SHORE}
\newcommand{\noboth}{\method{}\,$-$\allowbreak\,actor prior\,$-$\allowbreak\,critic weighting}

\title{\LARGE \bf
SHORE: Stability-Hybrid On-Policy Reinforcement Learning \\
for Online Bin Packing
}

\author{Chang Shi$^{1}$ and Ann Majewicz Fey$^{1}$ % <-this % stops a space
\thanks{$^{1}$University of Texas at Austin, Austin, TX 78712, USA}
\thanks{Correspondence to: Chang Shi \textless changshi@utexas.edu\textgreater }
}

\begin{document}

\maketitle
\thispagestyle{empty}
\pagestyle{empty}

%%%%%%%%%%%%%%%%%%%%%%%%%%%%%%%%%%%%%%%%%%%%%%%%%%%%%%%%%%%%%%%%%%%%%%%%%%%%%%%%

\begin{abstract}
Online 3D bin packing with a processing buffer requires choosing which box to pack, its orientation, and its placement while keeping the stack stable under unknown future arrivals. A collapse ends the episode and forfeits the remaining volume. Existing methods use geometric rules or hard stability masks, limiting how risk can influence the joint choice of box, orientation, and placement. We propose Stability-Hybrid On-Policy Reinforcement Learning (\method{}), which combines a learned stability probability of a placement with an on-policy estimate of its return conditional on stability. A 3D U-Net learns stability from executed outcomes and simulated alternatives, revealing outcomes of actions underrepresented in early rollouts. A discrete on-policy actor--critic uses stability predictions in both its action values and policy prior, with gates that limit the effect of an untrained or changing model. In deployment, \method{} does not require additional search or simulation overhead. In Isaac Gym, \method{} reaches 73\% space utilization by 0.4M environment steps, while the same learner without stability modeling reaches that level at about 1M. On the expanded 60-box set, \method{} achieves 0.734 space utilization versus 0.588 for the GOPT state-of-the-art baseline, with a higher placement-failure frequency (1.03\% vs. 0.63\%), trading more risk for greater packed volume.  The results show faster learning and higher utilization on the more demanding realistic box arrival set. Ablations suggest that simulated alternatives provide useful early supervision, while executed outcomes alone sustain later performance. We also demonstrate the method on a real robot to verify the simplicity of integrating the method to real bin packing pipelines.

\end{abstract}

\begin{figure*}[!t]
    \centering
    % trim (in px of the 1983x793 PNG): blank background above the robot and below the cart
    \includegraphics[width=0.9\linewidth, trim=0 61 0 24, clip]{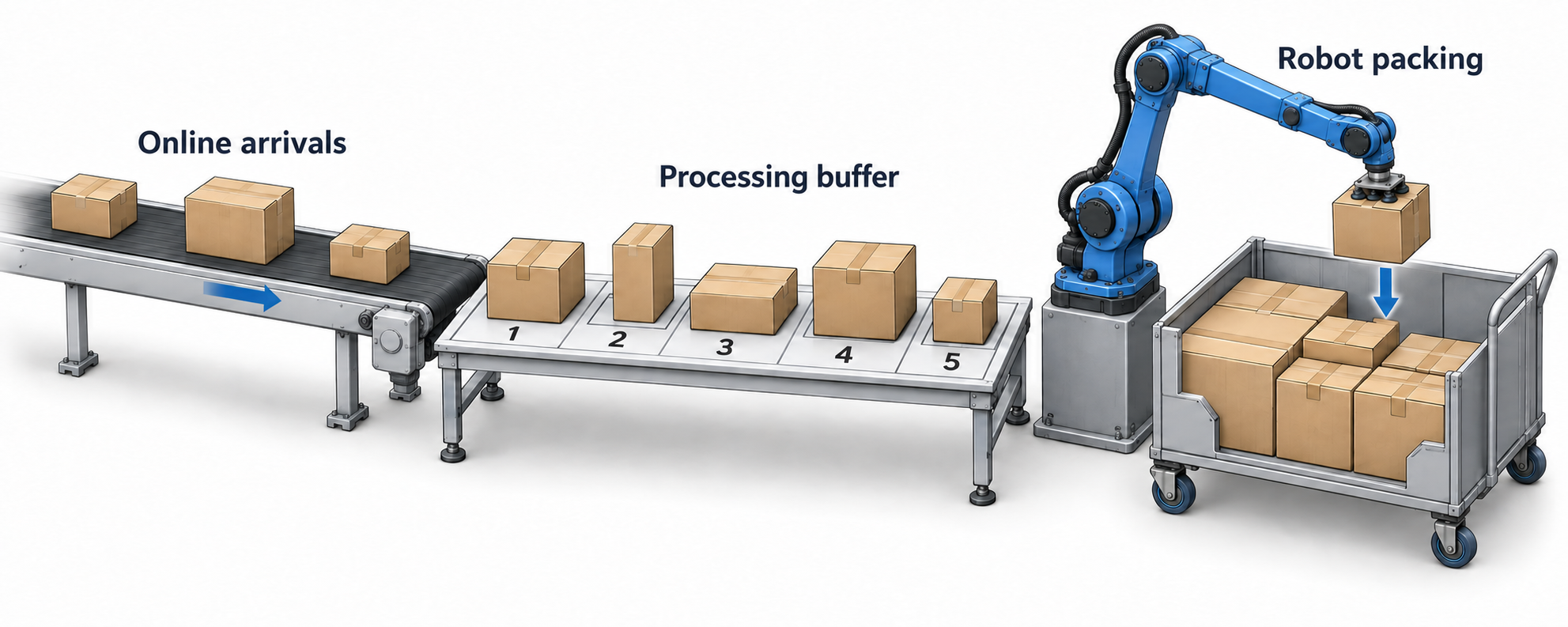}
    \caption{Illustration of the online bin packing setting with a finite processing buffer. The packing agent selects a box from the buffer, its orientation, and its placement while future arrivals remain unknown.}
    \label{fig:Problem illustration}
\end{figure*}

%%%%%%%%%%%%%%%%%%%%%%%%%%%%%%%%%%%%%%%%%%%%%%%%%%%%%%%%%%%%%%%%%%%%%%%%%%%%%%%%

\section{Introduction}
Online 3D bin packing requires selecting and placing boxes as they arrive, without knowing the complete arrival sequence. In warehouse settings, a small processing buffer allows for limited reordering prior to placement (Fig.~\ref{fig:Problem illustration}). The objective is to pack densely while keeping the stack stable, which depends on the box dimensions, physics, and the order, orientation, and location of placement.

Physical stability depends on more than the visible top surface. Supporting material properties and the arrangement of multiple supporting boxes can affect a placement's outcome. One box collapse can leave the bin disorganized and hurt the long-term packing compactness. Therefore, there have been works trying to integrate physics-aware stability information into packing decision making.

Reinforcement learning (RL) packers usually handle stability with an action mask. Most masks are computed using geometric or physics-based rules on the height map or the stacking structure \cite{zhao2022learning, xiong2024gopt}. Instead, \cite{zhang2025physics} and \cite{wu2024efficient} learn the stability mask through simulation outcomes together with a Proximal Policy Optimization (PPO) policy~\cite{schulman2017proximal}, but apply the masks as hard filters on the placement of a box the policy has already chosen. Stability therefore cannot influence which box to pack, leading to a policy that is not jointly optimal. Moreover, a placement is either allowed or forbidden, with no trade-off between its risk and the volume it adds. This motivates designing a more advanced joint policy in which stability awareness can directly affect selection among boxes, orientations, and placements in the long run.

We propose \method{}, which separates the prediction of the outcome of the current placement from the value of the subsequent packing, but intelligently fuses the stability information back to guide the long-term joint policy learning. A 3D U-Net reads bin occupancy and the oriented box properties to estimate whether the whole stack satisfies the stability test after the placement. A separate action-conditioned critic learns the return if the current placement holds, including its immediate reward and subsequent packing, from on-policy rollouts. The stability estimate modulates this value and provides a soft prior to the joint actions of the actor (Fig.~\ref{fig:framework}). In this on-policy actor--critic method, the critic scores every valid (box, orientation, placement) triple, and the actor follows the exact expected gradient over all of them, so the stability estimate shapes which box is packed and how it is turned, not only where it goes. At inference, \method{} chooses each placement with one pass of the actor and of the stability network.

In short, our contributions are:
\begin{itemize}
    \item A physics-aware RL strategy that uses learned placement-stability probabilities as continuous weights on action values for policy improvement over joint box--orientation--placement actions, so that the packing agent can trade the risk of an unstable placement against the volume it adds.
    \item A highly parallel, Isaac Gym-based bin packing environment that enables fast, large-scale on-policy RL training.
    \item An empirical study of the efficiency of this method on a realistic box distribution, together with an initial real-robot deployment test.
\end{itemize}

\section{Related Work}
\textbf{Online 3D Bin Packing.} This problem setting involves making packing decisions without knowledge of incoming items, leading to unique challenges in achieving optimality and efficiency. The early efforts on this problem are mostly studies about developing adaptive heuristics and algorithms that can handle the sequential nature of item arrivals. For example, \cite{karabulut2004hybrid} and \cite{ha2017online} utilize deepest-bottom-left (DBL) heuristics, while \cite{wang2019stable} introduces Heightmap-Minimization heuristics.

Recent work in this area has moved beyond classical heuristics to incorporate deep learning, physics constraints, and more generalizable models. \cite{verma2020generalized} use DRL to solve the specific setting with an arbitrary number of bins and any bin size, and show RL + search heuristics outperform traditional online heuristics across a range of instances. \cite{zhao2021online} tackle the problem with constrained DRL, framing the problem as a constrained Markov decision process under limited lookahead and enforcing physical stability, and show strong improvements over previous heuristics. \cite{yang2021packerbot}  designs a complete robot pipeline to narrow the gap of heuristic DRL framework between simulation and real world. More recently, \cite{xiong2024gopt} introduces a DRL model that not only places items but also generalizes to bins of varying sizes via a transformer architecture that fuses item and bin subspace features. \cite{zhou2024efficient} integrate a physics heuristic with object rearrangement and stable placement into a DRL framework to reduce wasted space and improve utilization. However, the design of good heuristics remains a challenge, which leads to the emergence of learning-based action masking models.

\textbf{Action masking models.} Action masking has emerged as a practical strategy to tackle the enormous, state-dependent action space in the online 3D bin packing setting. \cite{zhao2021learning} propose a feasibility-prediction module that learns a binary ``feasibility mask'' over placement choices and uses that mask to modulate actor probabilities during training, which substantially speeds learning and enforces collision/stability constraints. Building on the same idea, \cite{puche2022online} allows custom action-mask definitions including reorientation and empty maximal spaces (EMSs), and compares masking strategies with search-based planners. 
\cite{zhao2022learning} instead compute the feasibility mask for the placement actions based on the stacking tree analysis. \cite{wu2024efficient} incorporate physics properties into the mask calculation to improve sample efficiency and reduce unsafe actions during exploration. \cite{xiong2024gopt} uses pairwise action-masking between EMSs and orientations to guarantee that the policy only samples pairs, further improving generalization in the online setting. \cite{zhang2025physics} learn the mask online from physics simulations, jointly with the policy. All of these masks are hard filters on placements. \method{} instead keeps a geometric mask only for fit and support, and uses a learned stability probability inside the action value, where it can be traded against packed volume and affects which box is packed.

\section{Preliminaries}
\label{sec:prelim}
We consider online 3D single-bin packing with a processing buffer of size $N$. Only the first $N$ unplaced boxes are visible, and placing one admits the next arrival. The full simulator state $\xi_t$ includes the physical configuration and the entire arrival queue. It evolves as an MDP $\mathcal{M}=(\mathcal{X},\mathcal{A},P,\bar r,\gamma)$, where $\mathcal{X}$ is the full state space, $\mathcal{A}$ the joint action space, $P$ the transition kernel, and $\bar r(\xi,a)=\mathbb{E}[r_t\mid\xi_t=\xi,a_t=a]$ the expected reward. We use $\gamma=1$ and maximize $J(\theta)=\mathbb{E}_{\pi}[\sum_t\gamma^t r_t]$, the expected packed-volume fraction. The policy receives only the observation $s_t$ defined below, so its decision problem is partially observable. Arrival order is exogenous; the agent controls only the reordering allowed by the buffer.

\subsection{State and Observation}
The policy observation is $s_t=(C_t,d_t)$: the voxelized bin configuration and the properties of the visible boxes. The unseen arrival queue and unresolved contact details belong to $\xi_t$, not $s_t$. We use $s$ as shorthand for this observation in the policy, critic and stability model. The bin has physical dimensions $(L_p,W_p,H_p)$ and is discretized into $(l_p,w_p,h_p)$ cells. $C_t$ records voxel densities, and its height map $\eta_t\in\{0,\dots,h_p\}^{l_p\times w_p}$ gives the top occupied cell of each column. The buffer $d_t$ contains each visible box's dimensions and density. Following GOPT \cite{xiong2024gopt}, these observations are encoded as two token sets (Fig.~\ref{fig:framework}):
\begin{itemize}
    \item \textbf{Item tokens.} There is one token for each pair $u=(b,j)$ of a visible box $b$ and an orientation $j\in\mathcal{O}$, where $\mathcal{O}$ contains the six axis-aligned orientations. Token $\mathbf{x}_u$ contains the oriented extents, expressed in voxel units and normalized by $(l_p,w_p,h_p)$, the similarly normalized unrotated extents, a heavy flag derived from density, a presence flag and a selectability flag. Duplicate tokens remain in the observation but are not selectable.
    \item \textbf{Placement tokens.} There are at most $K$ empty maximal spaces (EMSs), each extending to the bin's maximum height. For EMS $k$, let $(x_1,y_1,z_1)$ and $(x_2,y_2,h_p)$ be its corners in voxel coordinates. Its token $\mathbf{y}_k$ contains these first five coordinates normalized by the corresponding grid dimensions, followed by support fractions $\rho_k^{\mathrm{flat}}$ and $\rho_k^{\mathrm{rigid}}$. These measure the part of its base supported at height $z_1$, and the part supported there by the bin floor or a heavy box, respectively. They are the only density-layout information provided to the actor and critic. EMSs are sorted by $(z_1,y_1,x_1)$ and truncated to $K$; unused slots are zero-padded and cannot be selected.
\end{itemize}

\subsection{Action}
\label{sec:action}
An action $a=(u,k)$ pairs an item token $u=(b,j)$ with an EMS $k$. Box $b$ is rotated to orientation $j$ and placed with its minimum corner at $(x_1,y_1)$ of EMS $k$, resting on the highest column under its footprint. Placements are therefore limited to EMS corners rather than every grid cell, and box order, orientation and placement are chosen jointly at every step. A rule-based mask $m(s,a)\in\{0,1\}$ admits a pair if and only if $u$ is selectable, the oriented box fits inside the EMS (and hence inside the bin and below $H_p$), and the placement passes GOPT's geometric support test \cite{xiong2024gopt}. The test requires that more than half of the footprint rests at the placement height, or that the footprint center lies within the convex hull of the supported cells. The valid action set is $\mathcal{A}(s)=\{a : m(s,a)=1\}$, with $|\mathcal{A}(s)| \le 6NK$.

\subsection{Reward Design}
The goal is to maximize space utilization while every placement stays stable. The reward at each step is the placed box's share of the bin volume:
\begin{equation}
    r_t=\chi_t\,\frac{V_{b_t}}{V_{\mathrm{bin}}},
    \label{eq:reward}
\end{equation}
where $b_t$ is the selected box, $V_{b_t}$ its volume, $V_{\mathrm{bin}}=L_pW_pH_p$, and $\chi_t\in\{0,1\}$ is the observed stability outcome: it is one exactly when Eq.~\eqref{eq:stability} holds. An unstable placement earns zero and ends the episode (Sec.~\ref{sec:episode_end}). With $\gamma=1$, the undiscounted reward sum is the stably packed volume fraction at termination, bounded by the total sequence volume ($0.691$ for the base set and $0.866$ for the expanded set). The reward of a successful placement depends only on its box volume; order, orientation and placement affect subsequent rewards through future fit and stability. Learners use the scaled reward $c_r r_t$, with $c_r=10$.

Near the end of an episode the buffer can hold fewer than $N$ boxes. Empty slots are marked absent and masked, so the policy cannot select them. No invalid-selection penalty is needed, unlike the penalty used in \cite{zhang2025physics}.

\subsection{Stability and Feasibility Criteria}
\label{sec:stability}
We make the two criteria used throughout the paper explicit.
\textbf{Stability (evaluated after placement).} A placement is executed by releasing the box $2$\,mm above its target position in the chosen orientation and simulating the scene forward for $\tau=80$\,ms. Let $\mathbf{p}_i^{\mathrm{tgt}},\mathbf{p}_i^{\mathrm{settled}}\in\mathbb{R}^3$ denote box $i$'s target and settled positions, and let $\mathcal{I}_t$ contain all boxes in the stack, including the new box. Stability requires
\begin{equation}
    \max_{i\in\mathcal{I}_t}\left\|\mathbf{p}_i^{\mathrm{settled}}-\mathbf{p}_i^{\mathrm{tgt}}\right\|_2\le\epsilon_s.
    \label{eq:stability}
\end{equation}
The threshold $\epsilon_s$ is a positional displacement tolerance; the test checks movement of both the new box and previously placed boxes.
\textbf{Feasibility (evaluated before placement).} Feasibility is the geometric mask $m(s,a)$ of the action definition. Apart from excluding duplicate tokens, it encodes only fit and GOPT's support test, both computed from the height map, so it does not predict Eq.~\eqref{eq:stability}, which depends on box densities and on the response of the whole stack. Stability is learned instead: \method{} trains a separate model of the probability that a placement satisfies Eq.~\eqref{eq:stability} (Sec.~\ref{sec:stab_model}) and uses it in both the critic and the policy.

\subsection{Episode End}
\label{sec:episode_end}
An episode ends in one of three ways: (i) \emph{completion}, when every box in the sequence has been placed; (ii) \emph{infeasibility}, when the method's feasibility filter admits no placement for the next decision; and (iii) \emph{instability}, when an executed placement violates Eq.~\eqref{eq:stability}. For \method{}, GOPT+ and GOPT, the feasibility filter is purely geometric: it is the rule-based mask $m(s,a)$ of Sec.~\ref{sec:action}, which admits a placement only if the oriented box fits inside the EMS and passes GOPT's support test, both computed from the height map. \method{} and GOPT+ apply it to every visible box in every orientation, so an infeasible ending means that no visible box fits anywhere with enough support; GOPT applies it only to the buffer head in its stored orientation. For OL Mask, the filter of the selected box also includes the learned stability mask, so its episodes can also end when the mask rejects every placement as unstable. An infeasible ending can therefore mean that no box fits in the remaining space, or that the filter rejects placements that would in fact be stable, and the evaluation does not separate the two.

\begin{figure*}[tb]
  \centering
  \includegraphics[width=\textwidth]{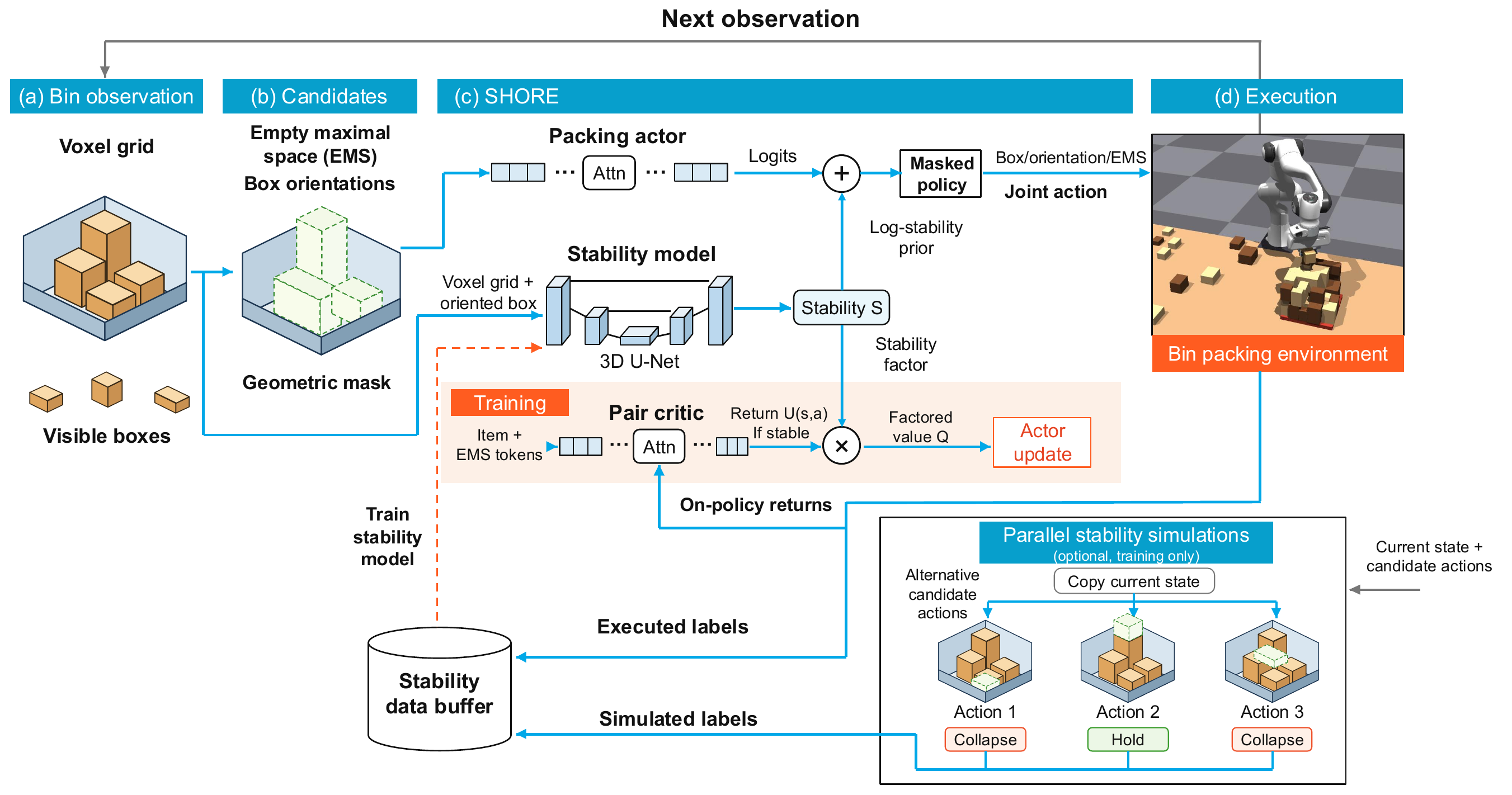}
  \caption{\textbf{SHORE framework.} The bin configuration $C_t$ and the buffer $d_t$ are encoded as EMS tokens and (box, orientation) item tokens, and the rule-based mask $m$ keeps the geometrically valid pairs. The actor, a Packing Transformer, scores every valid pair and samples the action $a=(u,k)$, which Isaac Gym executes and adjudicates by Eq.~\eqref{eq:stability}. A second Packing Transformer, the critic, values every pair in one pass, and a discretized REPPO learner updates both from on-policy rollouts (Sec.~\ref{sec:reppo}).}
  \label{fig:framework}
\end{figure*}

\section{Method}
\label{method}
% Implementation audit: ChangShiRaine/TP-PPO, branch holdq,
% commit 75b0fb29e18a7da0c722c6cd1a29705e0fa172a5.
% Core equations: holdQ/train_holdq.py; labels/model: holdQ/stability.py;
% actor/critic: reppo_tpt.py; placement physics: tpt_core.py and env_isaac_batched.py.
% Code b_prior -> w_pi, wf -> w_Q; exp(actor.log_lagrange) -> beta.
% The separate CLI --beta scales the prior and equals 1 in the reported runs.

\method{} trains an actor--critic over the joint action $a=(u,k)$ with a discretized version of REPPO \cite{voelcker2026relative} (Secs.~\ref{sec:networks}--\ref{sec:reppo}) and adds a learned estimate $S(s,a)$ of the probability that placement $a$ holds, i.e., satisfies Eq.~\eqref{eq:stability} (Secs.~\ref{sec:stab_model}--\ref{sec:stab_labels}). $S$ enters learning in two places: it weights the critic's value (\emph{critic weighting}, Sec.~\ref{sec:factored}) and it acts as a prior on the actor (\emph{actor prior}, Sec.~\ref{sec:prior}). Two safeguards keep an untrained or changing $S$ from disturbing the learner (Sec.~\ref{sec:safeguards}). Fig.~\ref{fig:framework} gives an overview, and Algorithm~\ref{alg:SHORE} lists one training iteration.

The stability model addresses a specific weakness of the plain learner. A collapse pays 0 and ends the episode, so a plain critic must learn, for every action, the product of two different quantities: the chance that the placement holds, which depends on the densities and support under the footprint, and the value of the packing that follows. The token observation shows EMS rectangles and two support fractions but not the interior of the stack, and collapses are rare, so the critic learns this product slowly and only for the actions the policy executes.

\subsection{Policy and Critic Networks}
\label{sec:networks}
The actor and the critic each have their own Packing Transformer backbone \cite{xiong2024gopt}. MLPs embed the item tokens $\mathbf{x}_u$ and EMS tokens $\mathbf{y}_k$ into $\mathbf{h}_u,\mathbf{e}_k\in\mathbb{R}^{128}$, and three encoder blocks apply self-attention within each token type and cross-attention between items and EMSs. Actor and critic parameters are denoted $\theta$ and $\phi$. With learned affine projections followed by LeakyReLU, denoted $F_1,F_2$, the base policy before adding stability information is
\begin{equation}
\begin{aligned}
    \ell_\theta(s,u,k)&=\langle F_1(\mathbf{h}_u),\,F_2(\mathbf{e}_k)\rangle,\\
    \pi_\theta^{\mathrm{base}}(a\mid s)&=\frac{m(s,a)\,e^{\ell_\theta(s,a)}}{\sum_{a'} m(s,a')\,e^{\ell_\theta(s,a')}}.
\end{aligned}
\label{eq:actor_net}
\end{equation}
The critic forms a pair feature $\mathbf{f}_\phi(s,u,k)=\mathrm{MLP}_\phi(P_h\mathbf{h}_u+P_e\mathbf{e}_k+P_s[\bar{\mathbf{h}};\bar{\mathbf{e}}])$, where $P_h,P_e,P_s$ are learned projections and the bars denote token means. Its head predicts a categorical distribution $p_\phi(i\mid s,a)$ on $n_b=151$ atoms $z_i$ spanning $[-40,40]$, with mean $Q_\phi(s,a)=\sum_i z_i p_\phi(i\mid s,a)$. The HL-Gauss encoding $\HLG(g)$ maps a scalar target $g$ to histogram-bin probabilities from $\mathcal{N}(g,\sigma_{\mathrm{HL}}^2)$, after clipping $g$ to the atom support and normalizing the retained bin masses, with $\sigma_{\mathrm{HL}}$ equal to 0.75 bin widths \cite{farebrother2024stop}. One forward pass scores all valid actions. Policy distributions are defined only for nonempty $\mathcal{A}(s)$; terminal states have zero continuation value.

\subsection{REPPO-D Backbone}
\label{sec:reppo}
We use the discrete REPPO formulation of \cite[Appendix~C]{voelcker2026relative}, denoted REPPO-D here. Its action-value expectation, entropy and KL divergence are evaluated exactly over the finite action set. In our setting, the sums cover the geometrically valid joint actions $\mathcal{A}(s)$, with the networks of Sec.~\ref{sec:networks} scoring every pair in one pass. This discrete update is inherited from REPPO, rather than a contribution of \method{}.

We retain REPPO's on-policy soft $\lambda$-returns, HL-Gauss critic loss with latent self-prediction, and KL-gated actor update with adaptive entropy temperature $\alpha$ and KL multiplier $\beta$; their definitions are given in \cite[Secs.~3.1--3.3 and Appendix~C]{voelcker2026relative}. The rollout-generating distribution is denoted $\pi_{\mathrm{old}}$. It remains fixed as the reference throughout the minibatch updates. Return targets use the scaled reward $c_r r_t$, zero continuation at terminal states, and value bootstrapping at rollout boundaries.

\method{} changes the action value, the actor's prior and entropy regularizer, and the entropy target, as detailed below. Its soft returns use the modified value in Eq.~\eqref{eq:shore_softv}, and its actor uses Eq.~\eqref{eq:shore_actor} inside the inherited KL gate. Setting $w_Q=w_\pi=0$ recovers our masked REPPO-D backbone, called \noboth{} in the experiments. Training settings are listed in Table~\ref{tab:hyperparameters}.

\subsection{Stability Model}
\label{sec:stab_model}
$S_\psi(s,a)\in[0,1]$, with parameters $\psi$, estimates the probability of satisfying Eq.~\eqref{eq:stability}. The voxel grid hides sub-cell offsets, velocities and contact states, so identical observed grids can have different placement outcomes. A probability supports expected-value estimation under this partial observability.

The network takes the voxel grid and one oriented box. Its input has six channels on the $l_p\times w_p\times h_p$ grid: occupancy, the voxel densities of $C_t$, and the box's length, width, height and density, each broadcast as a constant channel. Extents are normalized by the grid dimensions, and densities by the maximum box density. A 3D U-Net \cite{cciccek20163d} with the architecture of the OL Mask network \cite{zhang2025physics} (8 base feature maps, 3 levels, group normalization) ends in a convolution that collapses the height axis, giving 16 feature channels on the $l_p\times w_p$ grid. Five heads, each a $3\times3$ convolution, a ReLU and a $1\times1$ convolution, map these features to logit maps $g_\psi^{(n)}$, $n=1,\dots,5$, with one logit per placement corner. For action $a=(u,k)$, each head reads the map computed for item token $u$ at the corner $(x_1,y_1)$ of EMS $k$:
\begin{equation}
\begin{aligned}
    p_{\psi,n}(s,a)&=\operatorname{sigmoid}\big(g_\psi^{(n)}(s,a)\big),\\
    S_\psi(s,a)&=\tfrac15\textstyle\sum_{n=1}^{5}p_{\psi,n}(s,a),\\
    \nu_\psi(s,a)&=h_{\mathrm{bin}}\big(S_\psi(s,a)\big)\\
    &\quad-\tfrac15\textstyle\sum_{n=1}^{5}h_{\mathrm{bin}}\big(p_{\psi,n}(s,a)\big),
\end{aligned}
\label{eq:stab}
\end{equation}
where $h_{\mathrm{bin}}(p)=-p\log p-(1-p)\log(1-p)$ is binary entropy; all logarithms are natural. Entropy evaluation clips probabilities to $[10^{-6},1-10^{-6}]$ for numerical stability. A learning copy has parameters $\psi$ and a separate acting copy has parameters $\bar\psi$. Below, $S=S_{\bar\psi}$ and $\nu=\nu_{\bar\psi}$ denote the acting predictions, held fixed during actor--critic updates (Sec.~\ref{sec:safeguards}). For numerical stability, the policy and its regularizer use $\widetilde S(s,a)=\max\{S(s,a),\epsilon_{\log}\}$, with $\epsilon_{\log}=10^{-6}$; critic weighting uses the unmodified $S$. The head disagreement $\nu$ (BALD \cite{houlsby2011bayesian}) estimates epistemic uncertainty: it is high when the heads disagree, whereas uniformly uncertain heads can have low disagreement. It is used only to choose which placements to label (Sec.~\ref{sec:stab_labels}). One network pass covers every EMS of an item token, and tokens with equal extents and density share a pass, so a state costs at most $6N=30$ passes. Invalid actions get $S=1$ and $\nu=0$.

A separate convolutional network can use local density and support information absent from the actor--critic tokens. It also isolates supervised updates on buffered labels from the critic's on-policy representation.

\subsection{Stability Labels}
\label{sec:stab_labels}
A label is the binary outcome of a placement, $y=1$ if it held and $y=0$ if it collapsed, stored with its state in a buffer $\mathcal{D}_S$. Recorded physical outcomes can be reused across policy updates; predictive accuracy is checked on recent held-out placements as the visited states change. They come from two sources, collected during the rollout phase of Algorithm~\ref{alg:SHORE}:
\begin{itemize}
    \item \textbf{Executed placements.} Every placement executed in a rollout is labeled by its real outcome, with label weight $\zeta=2$. A random 10\% of executed outcomes go to a holdout set, excluded from supervised training and used to assess the learning copy $S_\psi$ before activation (Sec.~\ref{sec:safeguards}).
    \item \textbf{Simulated placements.} At every step, up to $B=B_s+B_r$ other valid placements per environment are tested in a copy of the environment and labeled by their simulated outcome, with label weight $\zeta=1$. They are never executed. The copies run in a pool of Isaac Gym environments alongside the rollout environments, so they settle in the same batched simulation. Up to $B_s=6$ placements are selected by descending score $\pi(a \mid s)^{\gamma_\pi}\nu(s,a)$, with $\gamma_\pi=0.5$ and at most two per item token, so that labels go to placements the policy might choose and $S$ is unsure about. Up to $B_r=2$ more are drawn uniformly without replacement from the remaining candidates, so that $S$ cannot keep confirming its own confident mistakes. Every eighth step, the executed placement is also simulated. Its simulated outcome is added to the training buffer and compared with the executed outcome to measure copy bias. Thus the holdout excludes executed outcomes, but can overlap simulated training queries at the same state--action pair.
\end{itemize}
The buffer keeps the labels of the most recent 40{,}000 visited states. After each rollout's actor--critic update phase, $S_\psi$ takes 64 Adam steps on a weighted binary cross-entropy with a bootstrap mask drawn once when each label is stored: independently, each head sees that label with probability $\tfrac12$. Keeping this mask fixed across reuse encourages head diversity:
\begin{equation}
    \mathcal{L}_S(\psi)=\frac{\sum_{i,n}\mu_{in}\,\zeta_i\,\mathrm{BCE}_{\mathrm{logit}}\big(y_i,g_\psi^{(n)}(s_i,a_i)\big)}{\max\{1,\sum_{i,n}\mu_{in}\,\zeta_i\}},
    \label{eq:bce}
\end{equation}
where $i$ indexes sampled labels, $\mu_{in}\sim\mathrm{Bernoulli}(\tfrac12)$ decides whether head $n$ sees label $i$, $\zeta_i$ is its source weight, and $\mathrm{BCE}_{\mathrm{logit}}$ is binary cross-entropy evaluated from logits. Executed labels weigh more because a simulated copy reports collapses that the real step does not have, mostly on tall stacks. With $B_s=B_r=0$ and the periodic copy check disabled, $S$ learns from executed placements only, and \method{} makes no simulator calls beyond the rollout itself.

\subsection{Stability-Factored Critic}
\label{sec:factored}
For a fixed policy and fixed return targets, let $G_t$ be the soft $\lambda$-return used to train the critic. It includes the scaled immediate reward and future soft values, but not the current action's entropy term. Thus $G_t=0$ on an immediate collapse. Define $S_{\mathrm{true}}(s,a)=\Pr(\chi_t=1\mid s_t=s,a_t=a)$ and $Q_G(s,a)=\mathbb{E}[G_t\mid s_t=s,a_t=a]$. Where $S_{\mathrm{true}}>0$, the law of total expectation gives $Q_G=S_{\mathrm{true}}\bar U_G$, with $\bar U_G=\mathbb{E}[G_t\mid s_t=s,a_t=a,\chi_t=1]$. \method{} writes the critic network's mean as $U_\phi$ and forms
\begin{equation}
    Q(s,a)=U_\phi(s,a)\,\big(1-w_Q+w_Q\,S(s,a)\big),
    \label{eq:qfactor}
\end{equation}
which replaces $Q_\phi$ in the soft value and the actor objective, with no gradient through $S$. The strength $w_Q\in[0,1]$ ramps from 0 to 1 (Sec.~\ref{sec:safeguards}). To move the critic target from the unconditional return toward the return conditional on survival, \method{} uses
\begin{equation}
\begin{aligned}
    \mathcal{L}_U=\;&\frac{\sum_{t\in\mathcal{B}_{\mathrm{AC}}}\omega_t\,
    \mathrm{CE}\big(\HLG(G_t),p_\phi(\cdot\mid s_t,a_t)\big)}
    {\max\{\epsilon_{\mathrm{loss}},\sum_{t\in\mathcal{B}_{\mathrm{AC}}}\omega_t\}}\\
    &+\mathcal{L}_{\mathrm{aux}},
\end{aligned}
    \label{eq:shore_critic}
\end{equation}
where $\mathcal{B}_{\mathrm{AC}}$ is an actor--critic minibatch of rollout transitions, $\epsilon_{\mathrm{loss}}=10^{-6}$ guards the denominator, $\mathrm{CE}$ is categorical cross-entropy, $\mathcal{L}_{\mathrm{aux}}$ is REPPO's unchanged latent self-prediction loss, and $\omega_t=1$ when $\chi_t=1$ and $1-w_Q$ otherwise. At a fixed $(s,a)$, the weighted scalar-target mean is $Q_G/(1-w_Q+w_QS_{\mathrm{true}})$ whenever the denominator is positive. If the critic represents this mean exactly, Eq.~\eqref{eq:qfactor} yields
\begin{equation}
    Q_{\mathrm{ideal}}(s,a)=Q_G(s,a)\,\frac{1-w_Q+w_Q\,S(s,a)}{1-w_Q+w_Q\,S_{\mathrm{true}}(s,a)}.
    \label{eq:ideal_factor}
\end{equation}
This equals $Q_G$ when $S=S_{\mathrm{true}}$ pointwise, or when $w_Q=0$. At $w_Q=1$, surviving samples train $U_\phi$ toward $\bar U_G$, which still includes the risk of later collapses. The identity explains the weighting design under exact mean fitting; finite data, function approximation, auxiliary training and HL-Gauss encoding can introduce additional error. In particular, low aggregate calibration error does not establish pointwise equality. If $w_Q=1$ and $S_{\mathrm{true}}=0$, all sample weights vanish and this conditional target is undefined.

This is where the per-action critic matters. A standard PPO critic predicts a state value; the per-action critic here permits stability to enter both value estimation and policy learning. Because $Q$ is a probability times a value, a slightly risky placement that adds much more volume can be worth more than a safe one, a trade that a hard mask cannot make.

\subsection{Stability Prior on the Policy}
\label{sec:prior}
$S$ also weights the policy directly. We write $\pi$ for the full policy, with dependence on $\theta,\bar\psi,w_\pi$ implicit:
\begin{equation}
    \pi(a \mid s)=\frac{m(s,a)\,e^{\ell_\theta(s,a)}\,\widetilde S(s,a)^{w_\pi}}{\sum_{a'}m(s,a')\,e^{\ell_\theta(s,a')}\,\widetilde S(s,a')^{w_\pi}},
    \label{eq:prior}
\end{equation}
with a prior strength $w_\pi\in[0,1]$ that ramps like $w_Q$. The prior is soft, unlike a hard mask. A valid placement that $S$ wrongly rates as unsafe retains nonzero sampling probability and can therefore receive a corrective executed label. When every candidate is risky, the policy renormalizes over the least risky ones instead of emptying the action set. The policy KL is computed on these full, stability-adjusted action probabilities (Sec.~\ref{sec:safeguards}).

Keeping REPPO-D's ordinary entropy regularizer would make the actor counteract this prior. The ordinary entropy-regularized objective pulls $\pi$ toward $\exp(Q/\alpha)$, so the actor would have to cancel the prior using voxel information it cannot see. \method{} instead regularizes relative to the prior, with sums over $a\in\mathcal{A}(s)$:
\begin{equation}
    \mathcal{J}_S(\theta;s)=\sum_{a}\pi(a \mid s)\Big[\alpha\log\frac{\pi(a \mid s)}{\widetilde S(s,a)^{w_\pi}}-Q(s,a)\Big].
    \label{eq:shore_actor}
\end{equation}
Up to a term that does not depend on $\theta$, $\mathcal{J}_S=\alpha\KL(\pi\,\|\,\pi_S)-\mathbb{E}_{\pi}[Q]$ with $\pi_S\propto m\,\widetilde S^{w_\pi}$. For fixed $Q,S,w_\pi$ and $\alpha>0$, its per-state minimizer over valid-action distributions is $\pi_{\mathrm{imp}}\propto \widetilde S^{w_\pi}\exp(Q/\alpha)$. Equation~\eqref{eq:prior} represents this distribution when $\ell_\theta(s,a)=Q(s,a)/\alpha+\kappa(s)$, where $\kappa(s)$ is any action-independent offset. Thus the ideal logits need not cancel the stability prior; neural approximation and the KL gate constrain the actual update. The soft value carries the same regularizer,
\begin{equation}
    V(s)=\sum_{a}\pi(a \mid s)\Big[Q(s,a)-\alpha\log\frac{\pi(a \mid s)}{\widetilde S(s,a)^{w_\pi}}\Big],
    \label{eq:shore_softv}
\end{equation}
with zero value at terminal states.

\textbf{Entropy target.} We retain REPPO-D's temperature update, which compares the policy's Shannon entropy $\mathcal{H}(\pi)=-\sum_{a\in\mathcal{A}(s)}\pi(a\mid s)\log\pi(a\mid s)$ with a target, but set
\begin{equation}
    \mathcal{H}_{\mathrm{tgt}}(s)=\tfrac12\mathcal{H}\big(\pi_S(\cdot\mid s)\big).
    \label{eq:shore_entropy}
\end{equation}
A target based only on the number of geometrically valid actions would demand the same exploration even when most candidates are predicted unstable. This target instead adapts exploration to the concentration of the stability prior. It reduces to $\tfrac12\log|\mathcal{A}(s)|$ when $w_\pi=0$, because the prior is then uniform over valid actions.

\subsection{Safeguards for a Learned Stability Model}
\label{sec:safeguards}
\textbf{Stability activation gate and ramp.} An untrained $S$ is noise, and as a prior it would bias early exploration. Both strengths therefore follow one schedule, $w_\pi=w_Q=w$, with
\begin{equation}
    w(n_{\mathrm{env}})=\mathrm{clip}\big((n_{\mathrm{env}}-n_{\mathrm{gate}})/N_{\mathrm{ramp}},\;0,\;1\big),
    \label{eq:ramp}
\end{equation}
where $n_{\mathrm{env}}$ counts total training environment steps, $N_{\mathrm{ramp}}=10^5$ is the ramp length, and $\mathrm{clip}(x,0,1)$ truncates $x$ to $[0,1]$. The gate step $n_{\mathrm{gate}}$ is the first training-iteration boundary at which the updated learning copy $1-S_\psi$ detects collapses among the held-out executed placements with an AUROC of at least 0.8, measured on at least 50 collapses. Once open, the gate stays open. Until it opens, $w=0$ and every equation above reduces to the learner of Sec.~\ref{sec:reppo}: $\pi\propto m\,e^{\ell_\theta}$, $Q=U_\phi$, all critic samples have weight 1, and $\mathcal{H}_{\mathrm{tgt}}=\tfrac12\log|\mathcal{A}(s)|$.

The activation gate protects early exploration from an uninformative stability model. Its AUROC criterion tests discrimination, not probability calibration; ECE is assessed separately in the experiments.

\textbf{What the policy KL gate protects.} The inherited REPPO-D gate limits actor updates relative to the rollout policy, where the on-policy critic has training support. In \method{}, it evaluates $\KL(\pi_{\mathrm{old}}\,\|\,\pi)$ over the full joint policy of Eq.~\eqref{eq:prior}, including the stability prior. The gate is applied per state: when its KL is below $\varepsilon_{\KL}$, the actor follows $\mathcal{J}_S$; otherwise, only the KL pullback is applied. These statewise losses are averaged over the minibatch. The acting stability model and the strengths $w_Q,w_\pi$ stay fixed throughout the rollout and its actor--critic updates. This gate controls policy change; it neither tests physical stability nor guarantees that a placement holds.

\textbf{Controlling stability-model updates.} Updating the acting stability model changes the policy even with the actor fixed, so it needs separate control. After training $S_\psi$, we interpolate $\bar\psi\leftarrow\bar\psi+\lambda_S(\psi-\bar\psi)$. A bisection search chooses a step $\lambda_S\in[0,1]$ whose induced policy KL, from the pre-interpolation policy to the candidate policy, stays within the mean bound $\delta_S$ on recent rollout states. The updated actor and the next rollout's prior strength are fixed during this check. If that strength is zero, the acting copy takes the learned weights directly. This bounds the measured policy drift caused by the model interpolation; it does not bound the separate change from advancing the strength ramp. It complements the actor's KL gate. The bound and state-sample size are given in Table~\ref{tab:hyperparameters}.

\begin{algorithm}[t]
\caption{\method{}: one training iteration}
\label{alg:SHORE}
\small
\begin{algorithmic}[1]
\Require actor $\theta$, critic $\phi$, learning/acting stability models $\psi,\bar\psi$, label buffer and holdout set
\State Set $w_Q,w_\pi$ from the activation ramp (Eq.~\eqref{eq:ramp})
\State Freeze $S_{\bar\psi}$ and both strengths for phases 1--2
\State Snapshot rollout policy $\pi_{\mathrm{old}}$ using Eq.~\eqref{eq:prior}
\Statex \textbf{1. Collect on-policy data and stability labels}
\For{each rollout step, in parallel environments}
    \State Sample $a_t\sim\pi_{\mathrm{old}}(\cdot\mid s_t)$
    \State Label candidate actions in copies of simulator state $\xi_t$ using the selection rule of Sec.~\ref{sec:stab_labels}
    \State Execute $a_t$; store transition and stability outcome
    \State Add simulated labels to the buffer; split executed labels between the buffer and holdout set
\EndFor
\Statex \textbf{2. Update the actor and critic}
\State Compute soft $\lambda$-returns using Eq.~\eqref{eq:shore_softv}
\For{each actor--critic minibatch update}
    \State Update $U_\phi$ with the weighted loss (Eq.~\eqref{eq:shore_critic})
    \State Update $\theta$ with Eq.~\eqref{eq:shore_actor} inside the REPPO-D KL gate relative to $\pi_{\mathrm{old}}$
    \State Update REPPO-D's dual variables, using the entropy target in Eq.~\eqref{eq:shore_entropy}
\EndFor
\Statex \textbf{3. Learn stability and prepare the next rollout}
\State Train $S_\psi$ on buffered labels with Eq.~\eqref{eq:bce}
\State Evaluate $S_\psi$ on held-out executed outcomes; record $n_{\mathrm{gate}}$ if the activation criterion first passes
\State Set the next rollout's strengths; move $S_{\bar\psi}$ toward $S_\psi$ subject to the stability-model KL check (Sec.~\ref{sec:safeguards})
\end{algorithmic}
\end{algorithm}

\subsection{Deployment}
\label{sec:deploy}
The critic, the label buffer and the simulator are used only during training. At deployment, \method{} computes the geometric mask, one actor pass and the stability network's passes for the visible items. It then executes $\arg\max_{a\in\mathcal{A}(s)}[\ell_\theta(s,a)+w_\pi\log\widetilde S(s,a)]$, the mode of Eq.~\eqref{eq:prior}. No search is run and no placement is simulated.

\section{Experiments}
The experiments answer three questions:

\begin{enumerate}
    \item Does explicit stability modeling accelerate policy learning and improve the trade-off between packed volume and collapse risk, beyond the benefit of buffer selection and rotation?
    \item Where must the stability estimate act: in the critic, in the policy, or in both?
    \item Do simulated stability labels help beyond the labels of executed placements, and is choosing them by uncertainty better than choosing them at random?
\end{enumerate}

\subsection{Baselines}
\label{sec:baselines}
All methods share the simulator, box sets, stability criterion (Eq.~\eqref{eq:stability}) and evaluation protocol (Sec.~\ref{sec:metrics}). The baselines keep their own action spaces and rewards. Except where stated, they place every box in its stored orientation, as in their original implementations, while \method{} and GOPT+ choose among six orientations.

\textbf{Masked-PPO variants.} \texttt{OL Mask} \cite{zhang2025physics} learns a stability mask online from physics simulations, jointly with a PPO policy. It is the strongest prior method in this setting and the closest to ours, since it also learns stability from physics. \texttt{Nomask} removes masking entirely. \texttt{Random Selection} randomly picks the box from the buffer with only the placement decided by the policy network; \texttt{Random Placement} places at random positions with only box selection determined by the policy.

\textbf{Learning-based state of the art.} \texttt{GOPT} \cite{xiong2024gopt} is a transformer-based DRL packer using Empty Maximal Spaces (EMSs --- the maximal axis-aligned empty boxes remaining in the bin) as placement candidates. We use the official implementation (Packing Transformer, EMS placement generator, and masked PPO) and swap only its geometric environment for ours, so that placements are executed and adjudicated by the same physics: unstable placements terminate episodes exactly as for the other methods. Since GOPT is a strict online packer without buffer selection, it packs the buffer head at each step. \texttt{GOPT+} extends GOPT to the action space of \method{}. The same Packing Transformer chooses any visible box in any of its six orientations, i.e., a joint (box, orientation, EMS) action, and is trained with GOPT's masked PPO \cite{schulman2017proximal}, with the clipped ratio applied separately to the (box, orientation) choice and to the EMS choice. Its rollout size, learning rate and number of updates are matched to \noboth{}, so GOPT+ has the actor network and the action space of \noboth{} and differs from it in the learner.

\begin{table}[h]
\caption{Box specifications in the simulation environment.}
\label{tab:box_specification}
\begin{center}
\begin{tabular}{cccc}
\toprule
Dimension (cm)      & Density (kg/m$^3$) & Rigidity & Counts\\
\midrule
6 × 6 × 4          & 100     & 0.5      & 10    \\
6 × 6 × 6          & 100     & 0.5      & 10    \\
6 × 6 × 6          & 5000    & 3        & 10    \\
8 × 6 × 6          & 5000    & 3        & 10    \\
\bottomrule
\end{tabular}
\end{center}
\end{table}

To assess sensitivity to box geometry, the expanded evaluation uses the 12 geometries in Tab.~\ref{tab:fba_boxes}, with five boxes per geometry (60 arrivals per episode), alongside the four-type base set. Density is assigned independently of geometry: each odd-numbered geometry contributes three light and two heavy boxes and each even-numbered geometry two light and three heavy, giving 30 light and 30 heavy boxes per episode. Light and heavy boxes use the two materials of the base set (100\,kg/m$^3$ with rigidity 0.5, and 5000\,kg/m$^3$ with rigidity 3; Tab.~\ref{tab:box_specification}). The 60 boxes fill 86.6\% of the bin volume, against 69.1\% for the base set, so the expanded set is both more heterogeneous and tighter. Every learned method is retrained from scratch on the expanded set with unchanged hyperparameters, and no policy is transferred between box sets. The training budget of each method is given with its results.
\begin{table}[t]
    \centering
    \caption{Box geometries for the FBA~\cite{carter2026fbabox}-inspired synthetic
    benchmark. Each geometry appears five times per episode.
    Density is assigned separately from geometry.}
    \label{tab:fba_boxes}
    \small
    \begin{tabular}{c c c c}
        \hline
        ID & Reference (in) & Simulation (cm) & Count \\
        \hline
        B01 & $8 \times 8 \times 6$
            & $4 \times 4 \times 3$ & 5 \\
        B02 & $10 \times 8 \times 6$
            & $5 \times 4 \times 3$ & 5 \\
        B03 & $12 \times 8 \times 6$
            & $6 \times 4 \times 3$ & 5 \\
        B04 & $12 \times 10 \times 6$
            & $6 \times 5 \times 3$ & 5 \\
        B05 & $12 \times 12 \times 8$
            & $6 \times 6 \times 4$ & 5 \\
        B06 & $14 \times 10 \times 8$
            & $7 \times 5 \times 4$ & 5 \\
        B07 & $16 \times 10 \times 6$
            & $8 \times 5 \times 3$ & 5 \\
        B08 & $16 \times 12 \times 8$
            & $8 \times 6 \times 4$ & 5 \\
        B09 & $16 \times 12 \times 12$
            & $8 \times 6 \times 6$ & 5 \\
        B10 & $18 \times 12 \times 8$
            & $9 \times 6 \times 4$ & 5 \\
        B11 & $18 \times 14 \times 10$
            & $9 \times 7 \times 5$ & 5 \\
        B12 & $20 \times 16 \times 12$
            & $10 \times 8 \times 6$ & 5 \\
        \hline
    \end{tabular}
\end{table}

\begin{table*}[t]
\caption{Performance comparison on the base box set. All methods are evaluated using the 1M trained checkpoint over 50 episodes with randomized arrival order and identical seeds. Episode metrics average over evaluation episodes. Space utilization and termination rates are dimensionless; episode length counts placed boxes and collapses per placement is in percent. Metrics are defined in Sec.~\ref{sec:metrics}. Completion rate is the fraction of episodes in which all boxes are placed. }
\label{tab:results_isaac}
\begin{center}
\begin{tabular}{lcccc}
\toprule
                & OL Mask \cite{zhang2025physics} & GOPT \cite{xiong2024gopt} & GOPT+
                & \textbf{\method{}(Ours)} \\
\midrule
Space utilization($\uparrow$)    & 0.495 & 0.661 & \textbf{0.683} & 0.657\\
Mean Episode Length($\uparrow$)  & 28.68 & 38.25 & \textbf{39.53} & 38.18\\
Unstable Rate($\downarrow$)      & 0.14 & 0.18 & \textbf{0.05} & 0.16\\
Collapses per placement (\%)($\downarrow$) & 0.49 & 0.48 & \textbf{0.12} & 0.42\\
Completion rate($\uparrow$) & 0.00 & 0.67 & \textbf{0.95} & 0.84\\
\bottomrule
\end{tabular}
\end{center}
\end{table*}

\begin{table*}[t]
\caption{Performance on the expanded box set (12 geometries in Tab.~\ref{tab:fba_boxes}, 60 arrivals per episode). All methods are evaluated using the 1M trained checkpoint over 50 episodes with randomized arrival order and identical seeds. Learned methods report mean $\pm$ standard deviation over three training seeds. No method completes the 60-box sequence, so the completion rate is 0 and omitted, and every episode that does not end in a collapse ends with no feasible placement (Sec.~\ref{sec:episode_end}). Collapses per placement is in percent (Sec.~\ref{sec:metrics}).}
\label{tab:results_fba}
\begin{center}
\setlength{\tabcolsep}{5pt}
\begin{tabular}{lcccc}
\toprule
                & OL Mask$^\ddagger$ \cite{zhang2025physics} & GOPT \cite{xiong2024gopt} & GOPT+ & \textbf{\method{}(Ours)} \\
\midrule
Space utilization($\uparrow$)    & 0.242$\pm$0.009 & 0.588$\pm$0.017 & 0.715$\pm$0.007 & \textbf{0.734}$\pm$0.006 \\
Mean Episode Length($\uparrow$)  & 16.45$\pm$1.07 & 41.08$\pm$1.06 & 50.55$\pm$0.69 & \textbf{50.83}$\pm$0.53 \\
Unstable Rate($\downarrow$)      & 0.99$\pm$0.01 & \textbf{0.26}$\pm$0.06 & 0.61$\pm$0.08 & 0.53$\pm$0.05 \\
Collapses per placement (\%)($\downarrow$) & 5.69$\pm$0.30 & \textbf{0.63}$\pm$0.15 & 1.20$\pm$0.17 & 1.03$\pm$0.10 \\
\bottomrule
\end{tabular}
\end{center}
\end{table*}

\subsection{Simulation Setup}
\label{sec:sim_setup}
We run all training and evaluation in Isaac Gym \cite{makoviychuk2021isaac} with the GPU-accelerated PhysX backend. A placement is executed as a release $2$\,mm above the target position in the selected orientation, without a full arm trajectory or added Gaussian position or rotation perturbations. This deliberate simplification keeps the property that matters for the decision problem, whether the stack holds, while abstracting away manipulator dynamics. A single box released on the bin settles onto its target with zero measurable drift.

For sim-to-sim transfer, we calibrated Isaac Gym's PhysX contact behavior against high-fidelity MuJoCo using paired placement replays (Appendix~\ref{sec:sim_calibration}). This aligns the faster simulator with MuJoCo's placement outcomes while retaining Isaac Gym's GPU-parallel capacity for large-scale RL training. Agreement is high but imperfect, so reported results remain specific to the calibrated Isaac Gym model.

In implementation, the bin is $25\times25$\,cm with a maximum stacking height of $20$\,cm, discretized into $25 \times 25 \times 20$ voxels of $1$\,cm for placement choice. The length of the processing buffer is set to 5. Our box configuration uses light and heavy densities of $100$ and $5000$\,kg/m$^3$, respectively, following the released implementation of \cite{zhang2025physics}. This density contrast makes density-aware ordering important. The experimental settings used for our comparisons are listed here rather than assumed to match all settings reported in the original paper. All remaining hyperparameters are listed in Tab.~\ref{tab:hyperparameters} (Appendix~\ref{sec:hyperparameters}).

With $N = 5$ and $K = 80$, each state has at most $6NK = 2{,}400$ joint actions, against $6Nl_pw_p = 18{,}750$ if every cell of the $25\times25$ grid were a placement choice. This action space is still large enough to make learning challenging.

Four types of boxes with different dimensions (length, width, and height), densities, and rigidities are used. While the density parameter controls how heavy the box is, the rigidity parameter controls the deformation degree of the box under contact forces; both impact placement stability. The detailed box specification is listed in Tab.~\ref{tab:box_specification}. The mixture of light deformable and heavy rigid boxes is what makes stability depend on more than geometry: for example, packing the light $6\times6\times4$ boxes first produces a soft top surface, and any heavy $8\times6\times6$ box arriving later must either be placed on that yielding surface (typically unstable) or terminate the episode. A policy that knows which surfaces will hold can choose its boxes and placements to keep rigid support available.

The total volume of all boxes in the base set is $0.691$ of the bin volume, which caps utilization even if every box is packed successfully. This near-saturation ceiling makes the base set a weak test of further packing gains. We therefore also evaluate on the expanded FBA set, whose broader range of 12 carton dimensions better represents varied fulfillment-box sizes and raises the all-boxes-packed ceiling to $0.866$. The expanded set offers more headroom for comparing packing strategies and makes box choice and orientation more consequential. The base set remains useful for checking whether the method's behavior transfers to a simpler, nearly saturated inventory.

\subsection{Evaluation Metrics}
\label{sec:metrics}
\emph{Space utilization} is the total stably packed box volume divided by bin volume. \emph{Mean episode length} counts boxes successfully placed before termination.

\emph{Completion rate} and \emph{unstable rate} report the fractions of episodes that end with every box placed and with a collapse (Sec.~\ref{sec:episode_end}). We do not report the share of infeasible endings as a metric: what it measures depends on each method's feasibility filter (Sec.~\ref{sec:episode_end}), so it is not comparable across methods. \emph{Collapses per placement} is the number of collapses divided by the number of placement attempts (placed boxes plus collapses) over a method's 50 evaluation episodes, in percent. An episode has at most one collapse. Thus, within a training seed, if $f_{\mathrm{col}}$ is the fraction of episodes ending in collapse and $\bar L$ is the mean number of successfully placed boxes, collapses per attempt is $f_{\mathrm{col}}/(\bar L+f_{\mathrm{col}})$ before conversion to percent. This normalizes failures by the number of executed attempts; it is not an independent outcome from episode length and unstable rate. It estimates the observed failure frequency under each policy's visited states, rather than risk on a common set of stacks. A low per-attempt frequency can still accompany frequent episode failures when many placements must succeed in succession.

\emph{Stability-model metrics.} The ablations (Sec.~\ref{sec:ablation}) evaluate the learning copy $S_\psi$ on held-out executed outcomes. These outcomes are excluded from supervised training, but simulated labels can cover the same state--action pairs (Sec.~\ref{sec:stab_labels}); the evaluation therefore does not establish generalization to entirely unseen placements. It also does not directly assess the acting copy $S_{\bar\psi}$.

\emph{AUROC}, the area under the receiver operating characteristic curve for $1-S_\psi$ detecting collapses, measures ranking: the probability that a collapsed placement receives a lower prediction than a successful one, with half credit for ties. A score of 0.5 is chance and 1 is perfect ranking. This metric opens the activation gate. \emph{Expected calibration error} (ECE) groups predictions into ten equal-width bins and averages the absolute difference between each bin's mean prediction and observed success frequency, weighted by bin size. Zero means perfect agreement at this bin resolution. Probability accuracy matters because the critic uses $S_{\bar\psi}$ multiplicatively, but low ECE does not prove the pointwise accuracy required by Eq.~\eqref{eq:ideal_factor}. The metrics are complementary: with rare collapses, a constant high prediction can have low ECE while ranking at chance; good ranking can coexist with pessimistic probabilities. Both metrics use the most recent held-out outcomes after each training iteration: about 4{,}000, including 37--66 collapses at the reported ablation checkpoint.

Although the box \emph{type} distribution is shared between training and evaluation (a realistic assumption for a warehouse with standardized box sizes), the \emph{arrival order} is independently randomized in every evaluation episode, so evaluation sequences are almost surely never seen during training; the combinatorial space of arrival orders is $\binom{40}{10,10,10,10} \approx 4.7\times10^{21}$. All metrics are averaged over 50 evaluation episodes with the same fixed episode seeds across all methods, using deterministic (argmax) actions and no search for the learned policies.

\subsection{Results}
\label{sec:results}

\textbf{Separating action flexibility from stability awareness.} The base-set comparison is close to its packing ceiling, while the expanded set tests the method in a more demanding regime: varied geometries constrain future fit, and material properties are independent of shape, so geometry alone cannot identify reliable support. At the base-set checkpoint, GOPT+ has higher utilization and completion than the single-seed SHORE run, despite SHORE's earlier rise in training reward. The curves therefore suggest faster early learning in this run, not a base-set evaluation advantage or a seed-robust difference. The expanded set provides the stronger test of whether modeling stability helps when box choice, orientation, and future fit interact.

\begin{figure}[!t]
    \centering
    \includegraphics[width=\linewidth]{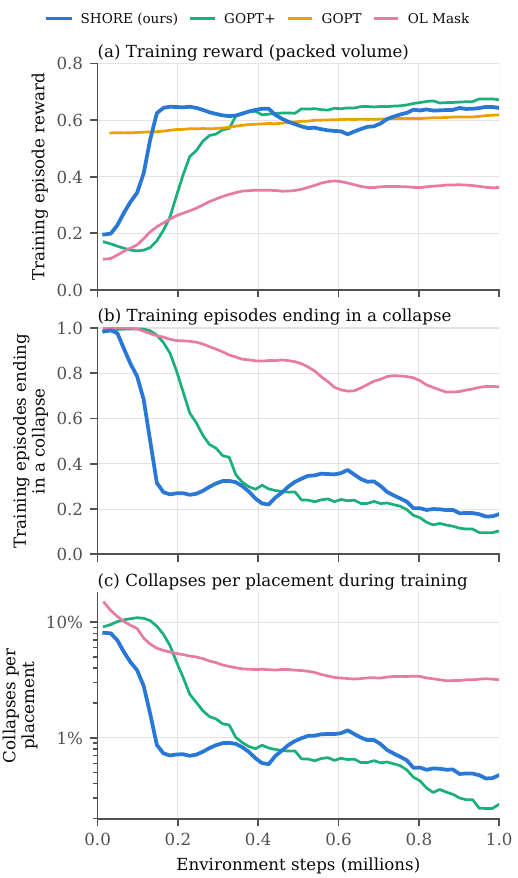}
    \caption{Training curves on the base box set for the methods of Table~\ref{tab:results_isaac}, up to 1M environment steps, computed from training episodes as in Fig.~\ref{fig:fba_curves}. (a)~Mean reward of training episodes (packed volume fraction). (b)~Fraction of training episodes that end in a collapse. (c)~Collapses per placement attempt (log scale). The original GOPT runs did not log how training episodes end, so GOPT appears only in (a). }
    \label{fig:base_curves}
\end{figure}

The comparison between GOPT and GOPT+ tests the value of broadening decisions beyond the buffer head and its fixed orientation. GOPT+ is our extension of the official GOPT baseline: it can choose a visible box, its orientation, and an EMS placement, so order, orientation, and location all affect the next packing state. Its stronger performance indicates that exposing these choices can improve packing on the expanded set and supports the importance of flexible, coordinated decisions. This is evidence for the value of the expanded decision space, though the comparison does not isolate action flexibility alone: GOPT+ also uses a different PPO update, with clipped ratios applied separately to item-orientation and placement choices.

Against this stronger diagnostic baseline, \method{} attains higher mean utilization while experiencing fewer collapses both per episode and per placement attempt, despite sharing its action space and actor backbone. The comparison therefore tests what stability-aware valuation and policy guidance add after the joint choices are available. To isolate stability modeling, the relevant control is REPPO-D, which shares the learner but removes both SHORE components. Its eventual utilization matches \method{} within training-seed variation. The evidence thus points to earlier acquisition and a better packing--risk balance, rather than a higher final utilization ceiling.

\begin{figure}[!t]
    \centering
    \includegraphics[width=\linewidth]{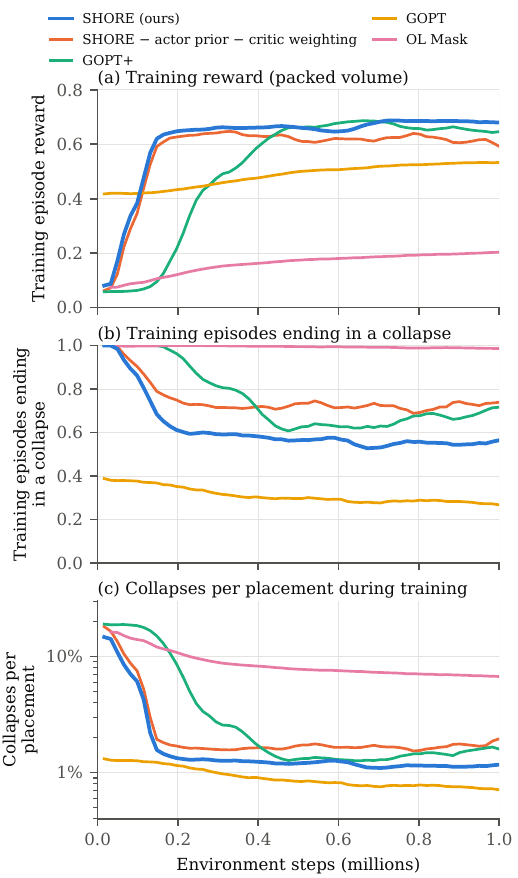}
    \caption{Training curves on the expanded (FBA) box set. (a)~Mean reward of training episodes (packed volume fraction), with actions sampled from each method's training policy. (b)~Fraction of training episodes that end in a collapse; the others end with no feasible placement. (c)~Collapses per placement attempt during training (log scale), pooled over the training episodes in each window. All three panels are computed from training episodes. A method whose episodes collapse long before the bin fills stays near 1 in (b) even as its collapses per placement fall in (c). The curves end at 1M steps, the checkpoint evaluated in Table~\ref{tab:results_fba}.}
    \label{fig:fba_curves}
\end{figure}

\textbf{Why stability supervision helps learning.} Figure~\ref{fig:fba_curves}(a) and the intermediate deterministic checkpoint evaluations show that \method{} reaches its high-utilization regime earlier than the learner without $S$. The advantage persists across intermediate checkpoints and appears under both sampled training actions and deterministic evaluation, so it is not merely a consequence of exploration noise in training returns. This pattern supports the hypothesis that separating immediate physical survival from continuation value makes learning easier. Without $S$, the critic must infer both effects from returns; a failed placement also removes the opportunity to observe later packing decisions. In \method{}, supervised placement outcomes train a dedicated model with access to the internal density layout, and that information immediately affects the values and probabilities of joint box--orientation--placement choices. The later catch-up of the base learner is consistent with $S$ supplying useful information sooner, while both ultimately optimize the same packed-volume objective. This learning advantage is measured in rollout environment steps.

\textbf{Stability without sacrificing useful packing.} Table~\ref{tab:results_fba} shows that \method{} packs more volume than GOPT+ with fewer failures per attempt, while their episode lengths are comparable. Its lower episode-collapse share thus accompanies a lower observed placement failure frequency, rather than simply fewer opportunities to fail. This supports using $S$ to guide productive joint actions. The two collapse metrics summarize the same failures at different scales; their agreement clarifies the packing--risk trade-off rather than providing independent statistical evidence.

Figure~\ref{fig:fba_curves}(c) shows how this trade-off develops during learning. After the initial improvement, \method{} sustains fewer collapses per attempt than the learner without $S$, alongside higher packed-volume reward in panel~(a). GOPT+ approaches its placement-level failure frequency later, consistent with explicit stability supervision accelerating the acquisition of useful physical constraints. OL Mask reveals why both collapse metrics are needed: its failures per attempt decline while its episode-collapse share stays near saturation. It sustains more successful placements before failure, but still fails before completion. Panel~(c) therefore reveals progress that panel~(b) obscures. These sampled training trajectories complement the deterministic evaluation comparison; they do not measure deployment-policy performance directly.

\textbf{Balancing stability against packed volume.} The expanded set shows two ways to get this trade-off wrong. GOPT is the conservative extreme. It has the lowest observed collapse frequency, yet packs much less volume than the methods that choose from the whole buffer. Its low collapse share is partly a consequence of stopping early: it places only the buffer head in its stored orientation, so most of its episodes end once that box has no admissible placement, after about 41 boxes, and each episode contains fewer placements that could fail. Its restricted actions also make each placement more conservative: it never turns a box onto a smaller face and never packs the last, tightest part of the bin. The training curves show the same regime throughout learning (Fig.~\ref{fig:fba_curves}): about 30\% of GOPT's training episodes end in a collapse, the fewest of any method, while its training reward stays near 0.5. GOPT+ is the aggressive counterpart. With buffer selection and rotation it packs more volume, but incurs a higher collapse rate. Because a collapse ends the episode and forfeits every box not yet packed, neither extreme is optimal: avoiding risk leaves space unused, and ignoring risk lets a single failed placement discard the volume the remaining boxes would have added. \method{}'s action value prices this trade-off directly, as the probability that a placement holds times the return if it holds. With the same action space and backbone as GOPT+, it attains higher utilization while experiencing fewer episode collapses and fewer failures per placement. The aim is thus not the lowest collapse rate, which GOPT shows can be bought by stopping early, but the most packed volume for the risk taken. OL Mask fails on both counts: its hard mask admits placements that do not hold, and 99\% of its episodes end in a collapse after about 16 boxes, although its different architecture, orientation restriction and training procedure prevent isolating soft versus hard masking as the cause.

No method completes the expanded sequences. For \method{}, fewer collapse endings accompany more geometrically infeasible endings, identifying preservation of usable space and support as a remaining challenge. Its soft prior does not remove geometrically valid actions; earlier choices can nevertheless leave no admissible placement. Infrequent placement failures can still interrupt a long sequence, while avoiding collapse alone does not preserve room for the remaining boxes. Normalization accounts for the number of attempts but not differences in the states visited; establishing lower risk on the same candidate placements would require a separate controlled comparison.

\begin{figure}[!t]
    \centering
    % In each panel the stack (825 px square) and the table crop (1232x731 px) have
    % equal heights at these widths (0.369 vs 0.623 x 731/1232 = 0.370).

    \vspace{3pt}
    \begin{subfigure}[t]{\linewidth}
        \includegraphics[width=0.369\linewidth]{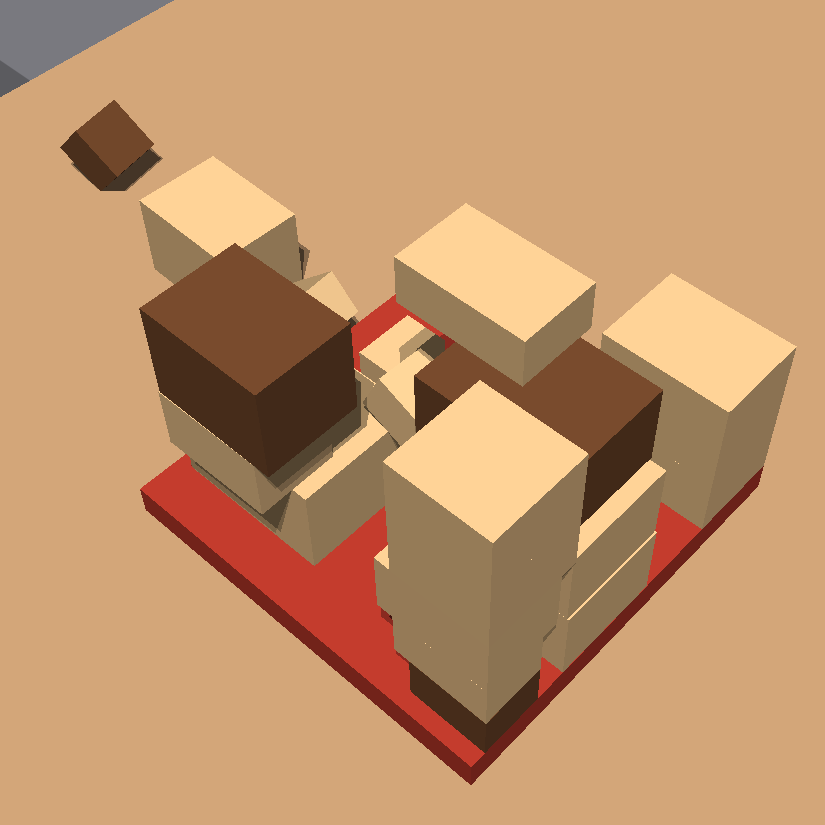}\hfill
        \includegraphics[width=0.623\linewidth]{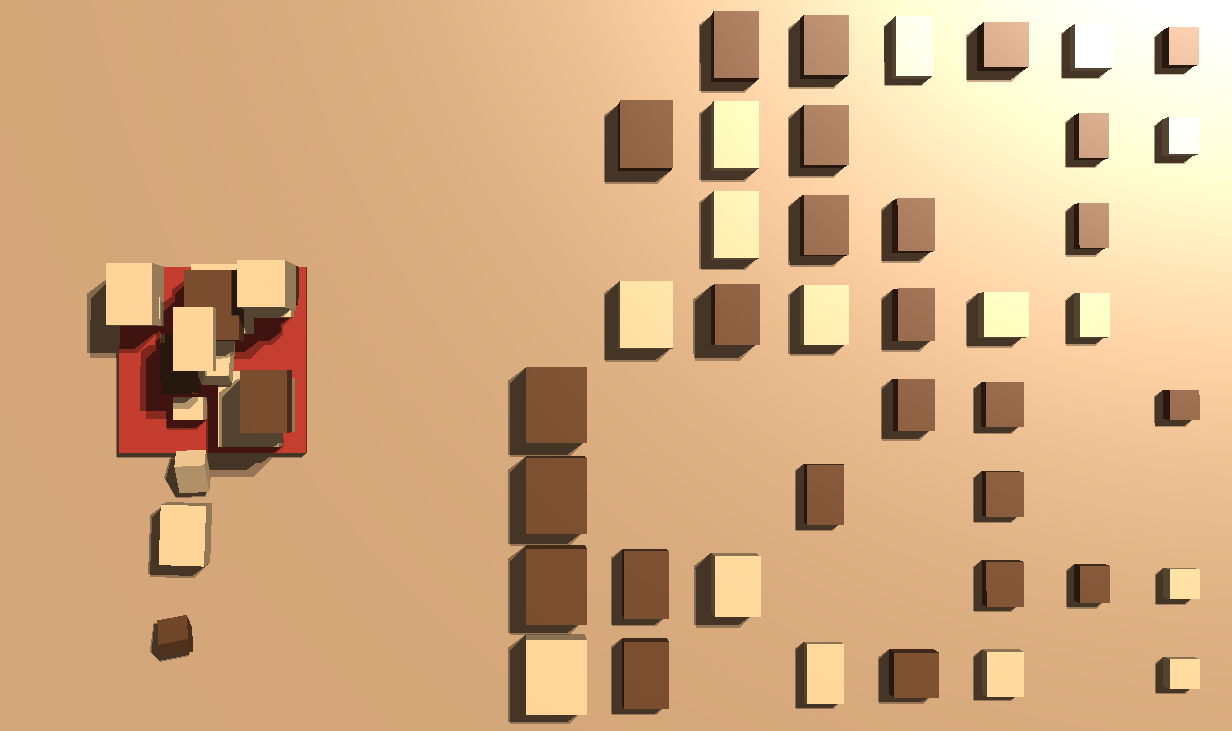}
        \caption{OL Mask: 40 boxes left on the table.}
        \label{fig:olmask_scene}
    \end{subfigure}

    \begin{subfigure}[t]{\linewidth}
        \includegraphics[width=0.369\linewidth]{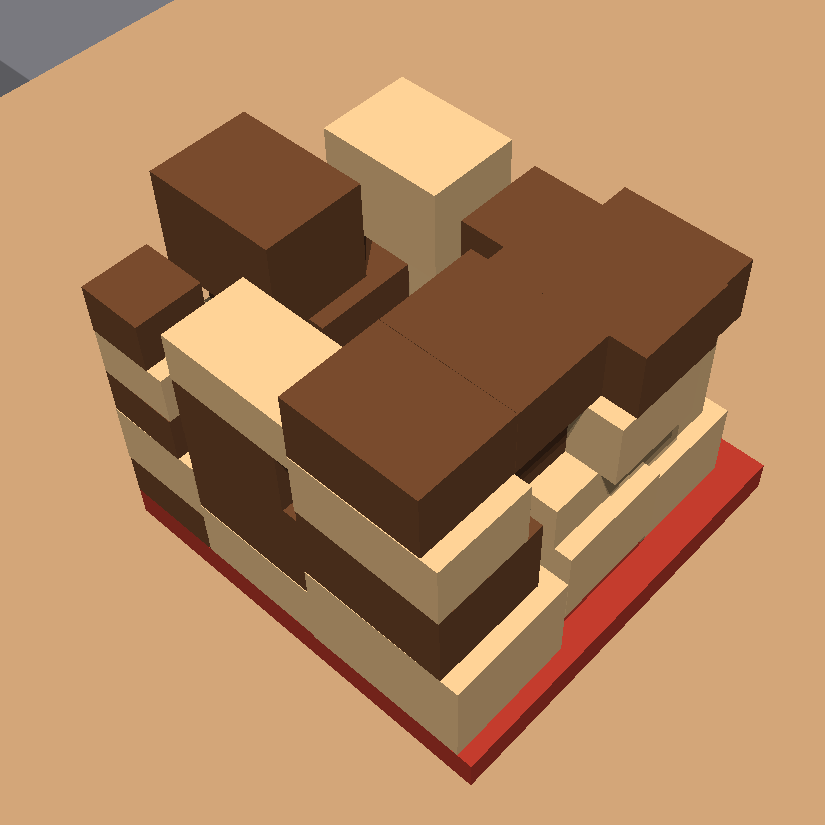}\hfill
        \includegraphics[width=0.623\linewidth]{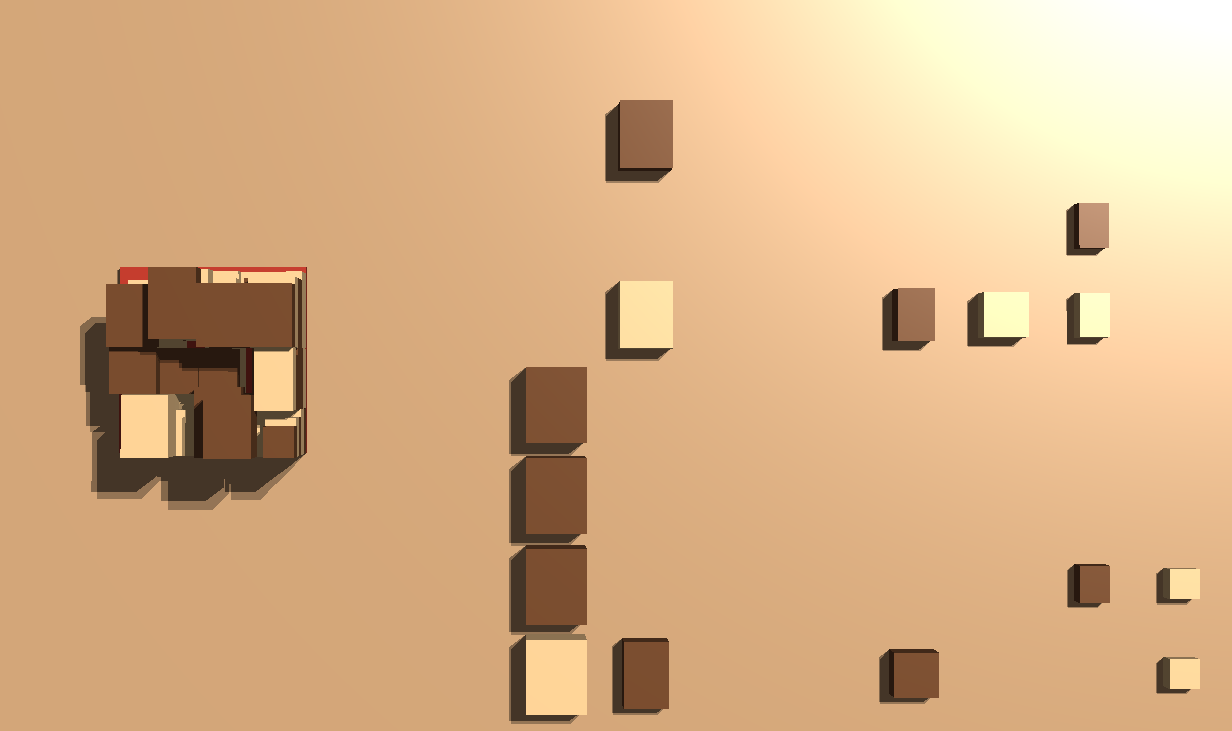}
        \caption{GOPT: 15 boxes left on the table.}
        \label{fig:gopt_scene}
    \end{subfigure}

    \vspace{3pt}
    \begin{subfigure}[t]{\linewidth}
        \includegraphics[width=0.369\linewidth]{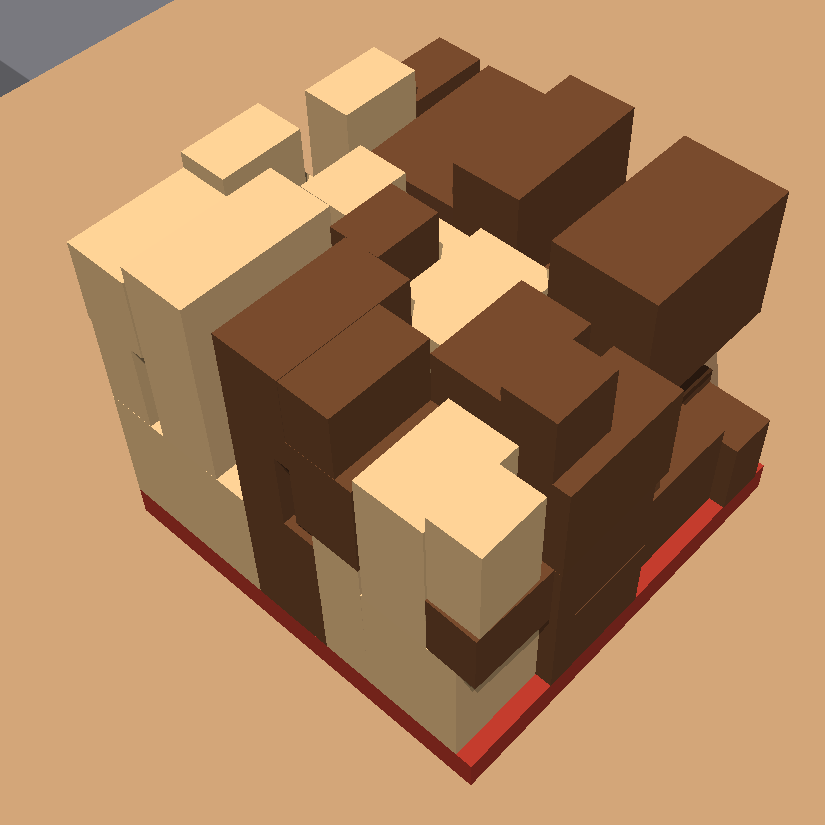}\hfill
        \includegraphics[width=0.623\linewidth]{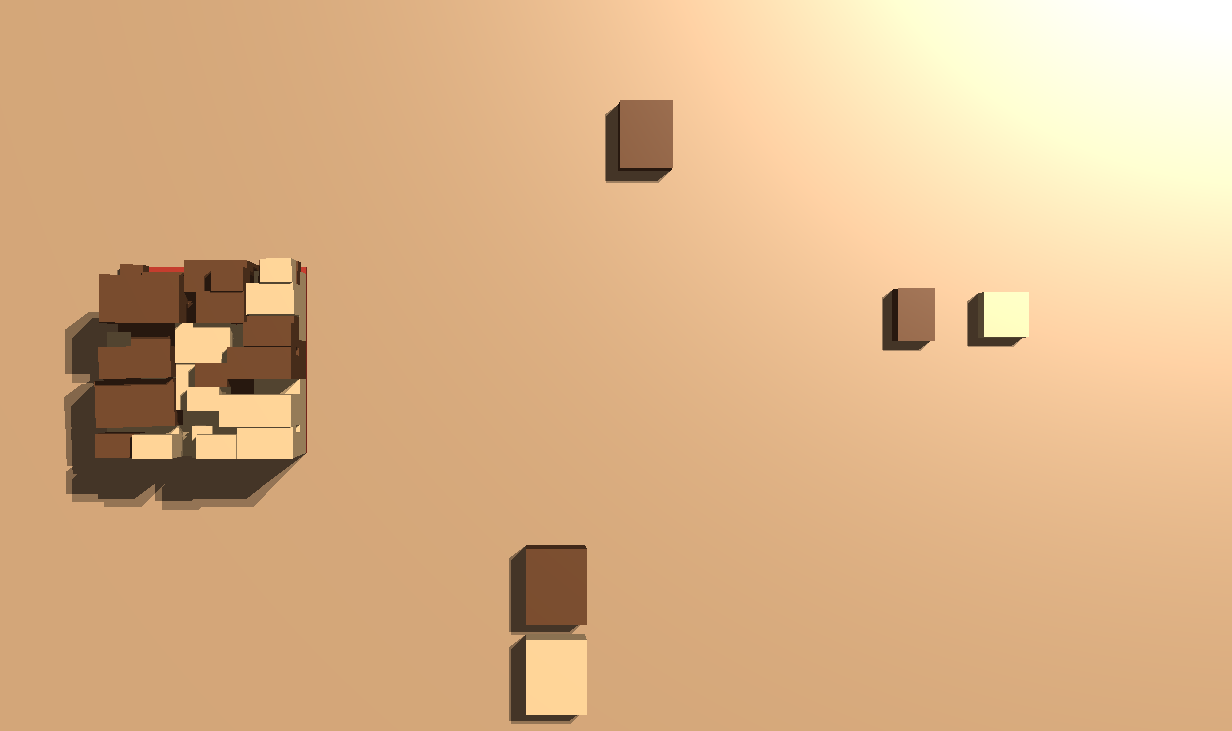}
        \caption{GOPT+: 5 boxes left on the table.}
        \label{fig:ppo_scene}
    \end{subfigure}

    \vspace{3pt}
    \begin{subfigure}[t]{\linewidth}
        \includegraphics[width=0.369\linewidth]{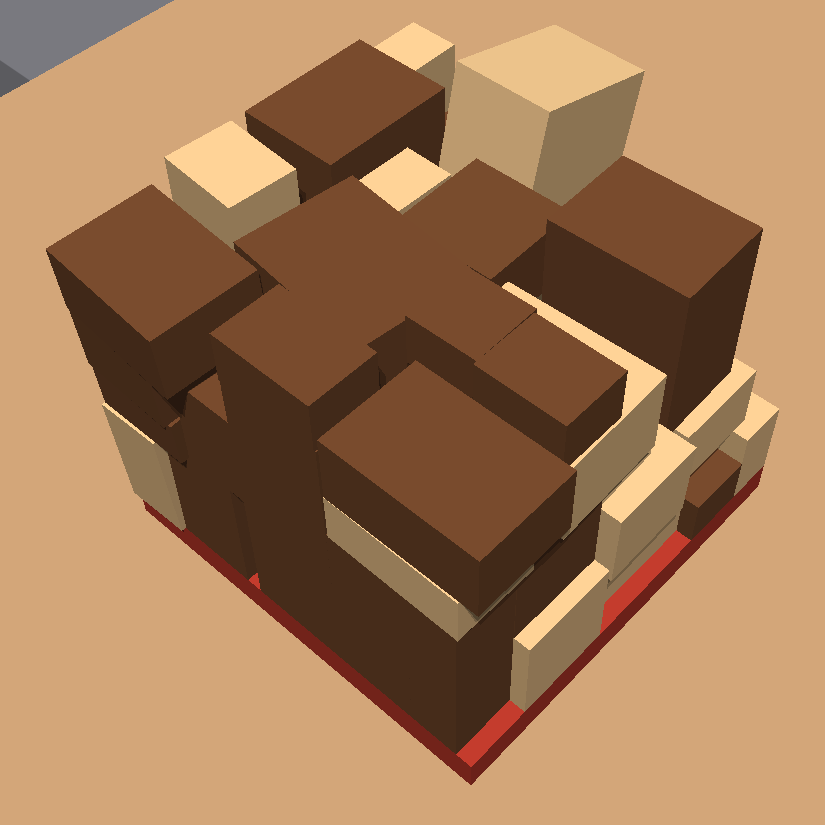}\hfill
        \includegraphics[width=0.623\linewidth]{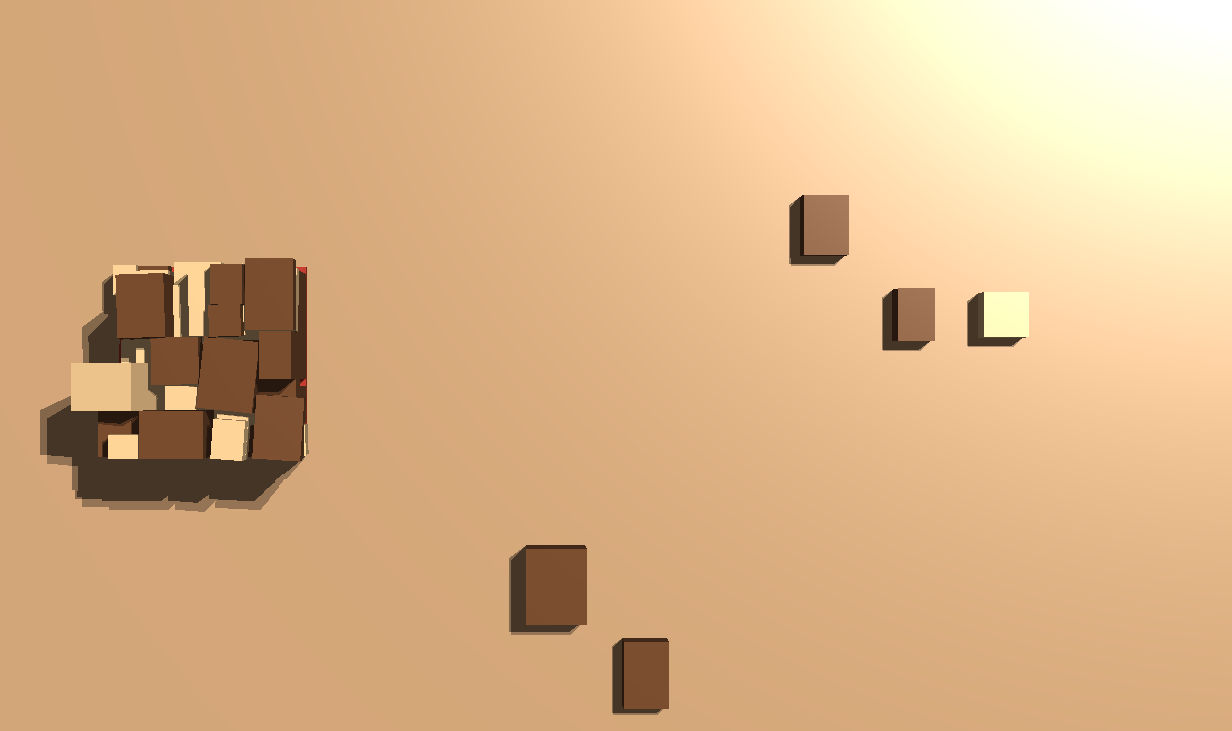}
        \caption{\method{}: 5 boxes left on the table.}
        \label{fig:shore_scene}
    \end{subfigure}
    \caption{An illustration of the scenes of one evaluation episode on the expanded box set, with the same box arrivals for all four methods, rendered in Isaac Gym after the episode ends. Each panel shows the final stack (left) and the table from above (right) with its stack and the boxes still waiting on the table. Pale boxes are light and dark boxes heavy; the red slab is the bin floor. OL Mask stopped when the last placement ended unstable and its collapse knocked boxes off the stack onto the table beside the bin.
    GOPT, GOPT stopped when on buffer boxes can find feasible placement. \method{} stopped when the last placement  fails  the  stability test, but the unstability did not leads to stack collapse.}
  \label{fig:visualization}
\end{figure}

Figure~\ref{fig:visualization} illustrates these distinct failure mechanisms under identical arrivals. We use the final checkpoint from the training seed with the median final score, and select the episode closest to the mean rewards of GOPT and \method{}. GOPT stops because the buffer head has no admissible placement, illustrating the restriction that buffer selection and rotation address. GOPT+ uses that flexibility to continue, but eventually exhausts admissible placements for every visible box. OL Mask instead admits a placement that destabilizes the stack, illustrating that passing a learned filter is not a physical guarantee. \method{} builds a substantially filled stack before a heavy placement fails; the remaining failure makes clear that its soft prior manages risk rather than eliminating it. The scene also shows why box count and utilization can disagree: \method{} packs fewer boxes than GOPT+ here but more volume. Evaluating the joint policy therefore requires the volume objective as well as the failure mode, rather than counting placements alone.

\begin{figure*}[t]
  \centering
  \includegraphics[width=\textwidth]{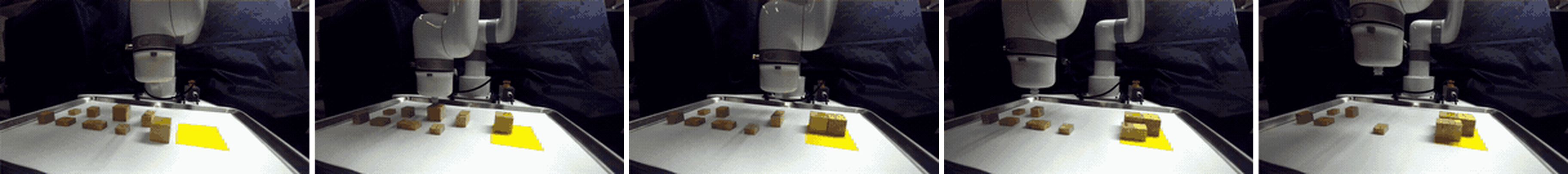}
  \caption{Real-robot packing with the UFactory Lite6. Key frames from the right camera view, in temporal order from left to right: the empty bin (yellow) and the stack after each of four successive placements. Boxes are picked from the table on the left of each frame.}
  \label{fig:real_robot}
\end{figure*}

\subsection{Ablations}
\label{sec:ablation}
The ablations change one part of \method{} at a time on the expanded set, with all other settings fixed, and fall into two groups. For the component ablation (Table~\ref{tab:ablation_modules}), \texttt{AP} (actor prior only) sets $w_Q=0$, \texttt{CW} (critic weighting only) sets $w_\pi=0$, and \texttt{REPPO-D} switches off both SHORE components ($w_\pi=w_Q=0$), leaving the discretized REPPO backbone of Sec.~\ref{sec:reppo}. The second group tests where the stability labels come from (Table~\ref{tab:ablation_labels}), with both components in place: executed placements only ($B_s=B_r=0$, no simulator calls beyond the rollout), one simulated label per step ($B_s=1$, $B_r=0$), eight uniformly random simulated labels ($B_s=0$, $B_r=8$), or the default eight, six chosen by $\pi^{\gamma_\pi}\nu$ and two at random ($B_s=6$, $B_r=2$). All variants are compared at the same checkpoint, 901{,}120 environment steps. Every run that trains a stability model was resumed from this checkpoint, and a resume empties the label buffer, so a later checkpoint would compare variants at different distances from a reset. Each variant is scored with the protocol of Sec.~\ref{sec:metrics}, and its stability model is also scored by AUROC (how well $S$ separates placements that collapse from placements that hold) and ECE (how closely $S$ matches the observed rate of holding), both defined in Sec.~\ref{sec:metrics}.

\begin{table}[t]
\caption{Ablation of where the stability estimate $S$ acts, on the expanded box set. Each variant removes the named components from \method{} (Sec.~\ref{sec:ablation}); the variants that keep $S$ use the default labels of Table~\ref{tab:ablation_labels}. AUROC (higher: $S$ ranks collapses below placements that hold more often) and expected calibration error (ECE; lower: the values of $S$ are closer to the observed rate of holding) score the checkpoint's $S$ on held-out executed placements (Sec.~\ref{sec:metrics}); they are n/a for \texttt{REPPO-D}, which has no stability model to score.}
\label{tab:ablation_modules}
\begin{center}
\footnotesize
\setlength{\tabcolsep}{2pt}
\begin{tabular}{lcccc}
\toprule
Variant & Score & Collapse & AUROC & ECE \\
 & ($\uparrow$) & ($\downarrow$) & ($\uparrow$) & ($\downarrow$) \\
\midrule
\method{} & \textbf{0.734} & \textbf{0.53} & 0.85 & \textbf{0.040}\\
\texttt{CW} & 0.708 & 0.60 & \textbf{0.86} & 0.051\\
\texttt{AP} & 0.699 & 0.58 & 0.82 & 0.042\\
\texttt{REPPO-D}$^\ast$ & 0.699 & 0.62 & n/a & n/a\\
\bottomrule
\end{tabular}
\end{center}
\end{table}

\begin{table}[t]
\caption{Ablation of the stability label source on the expanded box set, with both the actor prior and the critic weighting in place. $B_s$ simulated labels per step are chosen by $\pi^{\gamma_\pi}\nu$ and $B_r$ uniformly at random; executed placements are always labeled.}
\label{tab:ablation_labels}
\begin{center}
\footnotesize
\setlength{\tabcolsep}{3pt}
\begin{tabular}{lccccc}
\toprule
Labels & $B_s$/$B_r$ & Score & Collapse & AUROC & ECE \\
 & & ($\uparrow$) & ($\downarrow$) & ($\uparrow$) & ($\downarrow$) \\
\midrule
\method{} (default) & 6/2 & \textbf{0.734} & 0.54 & \textbf{0.85} & 0.040 \\
Executed only & 0/0 & 0.731 & \textbf{0.52} & 0.77 & 0.009 \\
1 simulated & 1/0 & 0.716 & \textbf{0.52} & 0.78 & \textbf{0.004} \\
8 random & 0/8 & 0.732 & \textbf{0.52} & 0.79 & 0.035 \\
\bottomrule
\end{tabular}
\end{center}
\end{table}

The component and label-source variants are exploratory single-seed comparisons, whereas \method{} and \noboth{} average multiple training seeds. Their small differences in packing or collapse share should therefore be treated as directional evidence. The reported metrics describe the learning copy, whereas decisions use the interpolated acting copy. Stability-model metrics also fluctuate across training iterations and are measured on each policy's own held-out executed outcomes, rather than a common candidate set. They diagnose behavior on the actions each policy encounters; they do not establish uniform accuracy across the joint action space.

\textbf{Complementary roles of value weighting and the action prior.} Table~\ref{tab:ablation_modules} supports complementary roles for the actor prior and critic weighting: retaining either alone does not recover the full method's packing performance. Critic weighting makes the value reflect the chance of losing both the current reward and the continuation, but its influence on action selection must pass through actor learning. The prior changes relative action probabilities directly, using voxel information unavailable to the actor, but by itself does not separate survival probability from the critic's return estimate conditional on survival. Their combination supplies both a physically informed action value and an immediate bias toward actions likely to hold. The observed ordering is consistent with this division of labor, although repeated seeds are needed to establish the interaction reliably.

The \texttt{CW} variant is especially informative: its stability model ranks outcomes at least as well as the full model at the reported checkpoint, yet its packing performance is lower. Prediction quality alone therefore does not explain the full method's advantage; how the estimate enters decision making matters. This is empirical support for the integration of $S$ into both modules, rather than evidence that a more accurate classifier automatically yields a better packer. The value factorization in Sec.~\ref{sec:factored} supplies the mathematical justification under pointwise accurate stability probabilities and exact mean fitting; these ablations test its practical usefulness with an imperfect learned model. They do not independently validate the stability activation gate or stability trust region, which are retained in the tested variants.

\textbf{What additional simulated labels contribute.} Table~\ref{tab:ablation_labels} shows that executed outcomes alone suffice to reach comparable packing performance at the ablation checkpoint. Together with the earlier checkpoint advantage of the default label scheme, this suggests that simulated alternatives mainly help acquire useful stability information before the rollout has supplied enough examples. They are not required to sustain the observed late-stage packing quality. Uncertainty-directed sampling also gives no demonstrated packing advantage over uniform sampling at this budget. The evidence therefore supports the stability model more strongly than it supports a particular acquisition rule or the necessity of extra simulator calls.

The ranking and calibration results explain why more labels need not improve control. The default simulated-label scheme yields stronger outcome ranking but poorer probability calibration than learning from executed outcomes alone. The recorded copy bias provides a plausible mechanism: simulated copies sometimes report collapse where the executed placement holds, pushing $S$ toward pessimistic probabilities. A ranking metric can improve despite this shift, while the multiplicative critic weighting depends on the probabilities themselves. Conversely, low ECE on mostly successful executed placements need not imply good discrimination of rare failures (Sec.~\ref{sec:metrics}). These findings support assessing both properties and caution against using either as a surrogate for packing performance. They also identify a concrete improvement path: reduce copy bias and assess label acquisition at matched simulation budgets, while retaining executed outcomes as the calibration reference.

\section{Real Robot Deployment}
We demonstrate the packing policy on a UFactory Lite6 arm with a vacuum gripper. The system uses a ZED 2i stereo camera and a hand-mounted camera. Fig.~\ref{fig:real_robot} shows one illustrated placement sequence. The yellow area serves as the packing bin. The sequence of cubes arrives in the order from right to left. This demonstration shows that our policy can be easily integrated into a robot packing pipeline.

\section{Conclusion and Future Work}
We presented \method{}, an on-policy reinforcement learning method for physics-aware online 3D bin packing, which incorporates learned stability information through a policy prior and continuous weighting of action-conditioned critic values. The experiments support faster acquisition of high-utilization policies than learning without explicit stability modeling. Against GOPT+, which shares the joint action space and actor backbone, \method{} achieves higher mean utilization with fewer collapses per episode and per placement attempt. The training curves further show that the stability model reduces failures per attempt relative to the same learner without it. These findings support using stability information in both action valuation and selection to improve the packing--risk trade-off; the eventual catch-up of the base learner limits claims of a higher final performance ceiling.

Meanwhile, the method is limited by its observation and candidate representations, approximate stability probabilities, and simulator fidelity. A collapse is attributed to the triggering placement even when earlier choices created the fragile structure; those earlier choices receive information only through returns, and the value network cannot see the raw voxel grid of the bin. Collapses per placement account for the number of attempts, but remain specific to the states each policy visits; earlier infeasible termination can prevent exposure to difficult late-stage placements. Stability metrics must therefore be interpreted together with utilization and termination modes. Besides, there are more sim-to-real factors that we did not take into account, such as contact-parameter mismatch, bin wall collisions, and robot reachability.

For future work, it is better to specifically distinguish exogenous factors like incoming box distribution from endogenous factors, and tackle the problem from the exogenous MDP perspective \cite{wan2024exploiting}. Techniques can include, but are not limited to, exogenous state decomposition~\cite{dietterich2018discovering} or training a predictive model of the exogenous factors. 

\bibliographystyle{IEEEtran}
\IEEEtriggercmd{\enlargethispage{2in}}
\bibliography{icra.bib}

\useRomanappendicesfalse
\appendices
\section{Simulation Calibration and Hyperparameters}
\label{sec:sim_calibration}\label{sec:hyperparameters}
\noindent\textbf{Sim-to-sim calibration against MuJoCo.} We used a system-identification-style paired-replay method: identical placement sequences were run in MuJoCo and Isaac Gym, and PhysX parameters were tuned to match per-placement stability verdicts. This increased agreement from 67.0\% before calibration to 94.8\% after calibration. MuJoCo served as the high-fidelity reference for sim-to-sim transfer; large-scale training and evaluation use Isaac Gym.

\par\medskip\noindent\textbf{Hyperparameters.} Tab.~\ref{tab:hyperparameters} lists the experiment settings and hyperparameters of \method{}. The upper block covers the environment and discretized REPPO learner (Sec.~\ref{sec:reppo}); the lower block covers the stability model and its labels.

\begin{table}[h]
  \caption{Experiment settings and hyperparameters grouped by function.}
  \label{tab:hyperparameters}
  \centering
  \footnotesize
  \setlength{\tabcolsep}{2pt}
  \begin{tabular}{lc}
    \toprule
    Parameter & Value \\
    \midrule
    \multicolumn{2}{l}{\textbf{Simulation}} \\
    Buffer size $N$ & $5$ \\
    Bin length and width $(L_p,W_p)$ & $(25,25)$ cm \\
    Maximum stacking height $H_p$ & $20$ cm \\
    Bin voxels $(l_p,w_p,h_p)$ & $(25,25,20)$ (1 cm bins) \\
    Added placement noise & None \\
    Vertical release offset & $2$ mm \\
    Simulation timestep / settling time $\tau$ & $2$ / $80$ ms \\
    Stability threshold $\epsilon_s$ & $2$ cm \\
    PhysX solver iterations (pos. / vel.) & $16$ / $4$ \\
    PhysX contact / rest offset & $2$ mm / $0$ \\
    PhysX maximum depenetration speed & $1$ m/s \\
    Simulation pool & $288$ environments \\
    \midrule
    \multicolumn{2}{l}{\textbf{Action space and REPPO-D learner}} \\
    EMS candidates $K$ & $80$ \\
    Discount $\gamma$ / $\lambda$-return $\lambda$ & $1$ / $0.95$ \\
    Reward scale $c_r$ & $10$ \\
    KL bound $\varepsilon_{\KL}$ & $0.1$ \\
    Initial entropy temperature $\alpha$ / KL multiplier $\beta$ & $0.01$ / $0.01$ \\
    Entropy target $\mathcal{H}_{\mathrm{tgt}}(s)$ & $\tfrac12\mathcal{H}(\pi_S)$ \\
    HL-Gauss atoms / support & $151$ / $[-40,40]$ \\
    Learning rate (AdamW) / gradient clip & $3\times10^{-4}$ / $0.5$ \\
    Rollout (environments $\times$ steps) & $32\times128$ \\
    Epochs $\times$ minibatches (size) & $4\times8$ ($512$) \\
    \midrule
    \multicolumn{2}{l}{\textbf{Stability model and labels}} \\
    Stability heads / U-Net maps, levels & $5$ / $8$, $3$ \\
    Simulated labels per step $B_s+B_r$ & $6+2$ \\
    Label selection exponent $\gamma_\pi$ & $0.5$ \\
    Label weight (executed / simulated) & $2$ / $1$ \\
    Holdout share of executed labels & $0.1$ \\
    Label buffer (states) & $40{,}000$ \\
    Stability updates per iteration (lr; batch) & $64$ ($10^{-3}$; $256$) \\
    Activation gate: AUROC / held-out collapses & $0.8$ / $50$ \\
    Stability ramp length $N_{\mathrm{ramp}}$ & $10^5$ env steps \\
    Stability trust region $\delta_S$ / states & $0.02$ / $256$ \\
    \bottomrule
  \end{tabular}
\end{table}

\end{document}